\documentclass[lettersize,journal]{IEEEtran}

\usepackage{amsmath}
\usepackage{amssymb} 
\usepackage[table]{xcolor}
\usepackage{booktabs} 
\usepackage{multirow}
\usepackage{svg}
\usepackage{url}
\usepackage{longtable}
\usepackage{pdflscape}
\usepackage{float}
\usepackage{wasysym}
\usepackage{bbm}
\usepackage{import}
\usepackage{comment}
\usepackage{algorithm2e}
\usepackage{caption}
\usepackage{subcaption}
\usepackage[normalem]{ulem}
\usepackage{tikz}
\usepackage{pgfplots}
\usepackage{pgfplotstable}
\pgfplotsset{width=10cm,compat=1.18}
\usetikzlibrary{plotmarks}
\usepgfplotslibrary{statistics}
\usetikzlibrary{pgfplots.groupplots}
\usetikzlibrary{patterns} 
\usepackage[pagebackref=false,breaklinks=true,letterpaper=true,colorlinks,bookmarks=false]{hyperref}
\hypersetup{
    linkcolor=gray,
    filecolor=black,
    urlcolor=gray,
    citecolor=black,
}

\definecolor{tm1}{RGB}{233,180,40}
\definecolor{tm2}{RGB}{167,121,204}  
\definecolor{tm3}{RGB}{0,159,230}   
\definecolor{ac}{RGB}{141, 236, 13}
\definecolor{ads}{RGB}{91, 185, 72}
\definecolor{ag}{RGB}{13, 236, 232}
\definecolor{epe}{RGB}{214, 150, 45}
\definecolor{es}{RGB}{214, 92, 45}

\graphicspath{ {./images/} }

\newcommand{\bbox}{\protect\raisebox{1pt}{\protect\tikz \protect\draw[black,fill=black] (1,1) circle (0.5ex);}}
\newcommand{\wbox}{\protect\raisebox{1pt}{\protect\tikz \protect\draw[black,fill=white] (1,1) circle (0.5ex);}}

\begin{document}

\title{\resizebox{\textwidth}{!}{Visual Affordance Prediction: Survey and Reproducibility}}

\author{Tommaso Apicella, Alessio Xompero, Andrea Cavallaro%
\thanks{Tommaso Apicella is with Istituto Italiano di Tecnologia, Italy (e-mail: tommaso.apicella@iit.it).}%
\thanks{Alessio Xompero is with the Centre for Intelligent Sensing, Queen Mary University of London, London E1 4NS, U.K. (e-mail:
a.xompero@qmul.ac.uk).}%
\thanks{Andrea Cavallaro is with Idiap Research Institute and EPFL, Switzerland (e-mail:
andrea.cavallaro@epfl.ch).}
}

\makeatletter
\def\ps@IEEEtitlepagestyle{
  \def\@oddfoot{\mycopyrightnotice}
  \def\@evenfoot{}
}
\def\mycopyrightnotice{
  {\footnotesize
  \begin{minipage}{\textwidth}
  \tiny
  \copyright\ 2025 IEEE. This work has been submitted to the IEEE for possible publication. Copyright may be transferred without notice, after which this version may no longer be accessible.
  \end{minipage}
  }
}

\markboth{Journal of \LaTeX\ Class Files,~Vol.~18, No.~9, October~2025}
{T. Apicella, A. Xompero, A. Cavallaro: Visual Affordance Prediction: Survey and Reproducibility}

\maketitle

\begin{abstract}
Affordances are the potential actions an agent can perform on an object, as observed by a camera. Visual affordance prediction is  formulated differently for tasks such as grasping detection, affordance classification, affordance segmentation, and hand pose estimation. This diversity in formulations leads to  inconsistent  definitions that prevent fair comparisons between methods. In this paper, we propose a unified formulation of visual affordance prediction by accounting for the complete information on the objects of interest and the interaction of the agent with the objects to accomplish a task. This unified  formulation allows us to comprehensively and systematically review disparate visual affordance works, highlighting strengths and limitations of both methods and datasets. We also discuss reproducibility issues, such as the unavailability of methods implementation and experimental setups details, making benchmarks for visual affordance prediction unfair and unreliable. To favour transparency, we introduce the Affordance Sheet, a document that details the solution, datasets, and validation of a method, supporting future reproducibility and fairness in the community.
\end{abstract}


\section{Introduction}
\label{sec:introduction}

\IEEEPARstart{A}{ffordances} are the potential actions that objects in the scene offer to an agent (i.e.~a human or a robot)~\cite{gibson1966senses}. Because of such a broad definition, the prediction of affordances is generally cast into different formulations, such as grasping detection, affordance classification, affordance segmentation, and hand-object interaction synthesis~\cite{lenz2015deep,pieropan2013functional,apicella2024segmenting,ye2023affordance}.
Each redefinition addresses a part of the affordance prediction problem. For example, affordance classification identifies what actions to perform; affordance detection and segmentation localizes which objects and what regions to interact with; and grasping detection predicts the object points to perform the interaction. 

Learning to perceive object affordances from visual data is challenging due to the varying appearance of objects based on the setting (e.g. single object on a tabletop or presence of clutter), the limited size of datasets, and the characteristics of the agent's hand influencing the interaction with objects. 
\textit{Environmental conditions}, such as illumination, background and clutter, camera viewpoint and distance, influence the target object affordance. For instance, occlusions caused by other objects in cluttered scenes~\cite{nguyen2017object,guo2023handal,corona2020ganhand} or by a human hand during a manipulation~\cite{xompero2020corsmal,sawatzky2017weakly} prevent the accurate perception of the target object's functional regions, potentially causing unintended collisions or unsafe interactions.
Moreover, different \textit{environments} (or contexts) imply different affordances for the same object. For example, a screwdriver can be used to insert or remove screws in a workshop through the graspable handle. In an environment where the object does not belong to (e.g. kitchen), the whole surface of the screwdriver becomes graspable to move it elsewhere. \textit{Object properties}, such as material (e.g. reflective), appearance (e.g. transparency or texture), and geometry (e.g. size or shape), also influence the observation of an object and the affordance. For example, concave shapes afford the holding of a content, and sharp regions afford cutting~\cite{myers2015affordance,do2018affordancenet,luo2022learning}. The physical characteristics of a \textit{hand} (human or robotic), such as size, degrees of freedom, and number of fingers, can influence the interaction with the object.  
A gripper with small fingers can grasp a wine glass from the stem, whereas a gripper with large fingers can grasp the glass bowl but not the stem.

\begin{figure}
    \centering
    \includegraphics[width=0.95\linewidth]{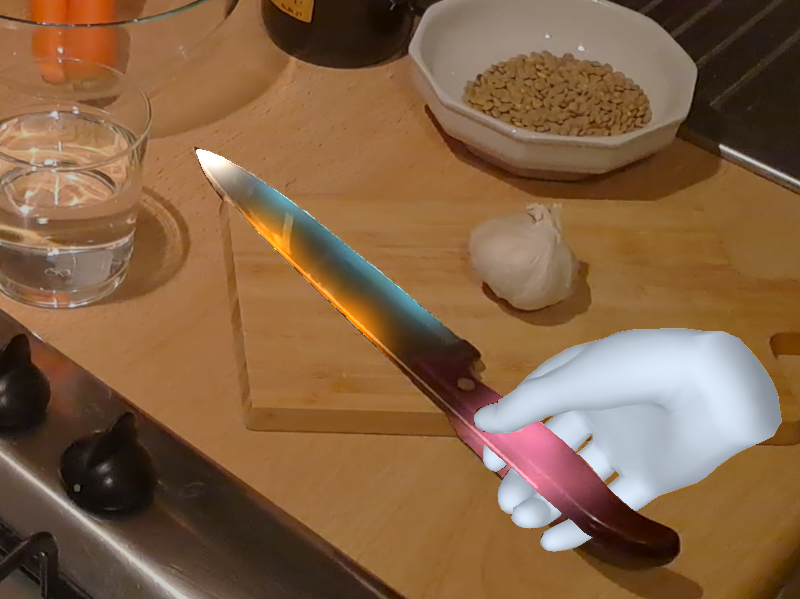}
    \caption{Visual affordance prediction in case of a knife: \textit{what} actions the agent performs, \textit{where} the hand interacts with the object (heat map), and \textit{how} the interaction is performed (hand pose for cutting). 
    Legend:
    \protect\raisebox{2pt}{\protect\tikz \protect\draw[tm2,line width=2] (0,0) -- (0.3,0);}~\textit{grasp},
    \protect\raisebox{2pt}{\protect\tikz \protect\draw[tm1,line width=2] (0,0) -- (0.3,0);}~\textit{slide},
    \protect\raisebox{2pt}{\protect\tikz \protect\draw[tm3,line width=2] (0,0) -- (0.3,0);}~\textit{cut},
    \protect\raisebox{1pt}{\protect\tikz \protect\draw[pattern color=black] (0,0) rectangle (0.3,0.1);}~\textit{pierce}.}
    \label{fig:whatwherehow}
    \vspace{-12pt}
\end{figure}

\begin{figure}
    \centering
    \includegraphics[width=\linewidth]{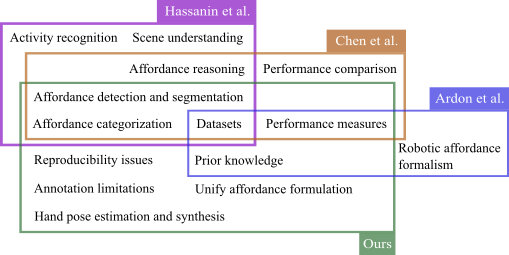}
    \caption{Comparison of topics discussed in affordance surveys: Hassanin et al.~\cite{hassanin2021visual}, Chen et al.~\cite{chen2023survey}, Ardón et al.~\cite{ardon2020affordances}. We propose an affordance formulation unifying previous redefinitions, we discuss the reproducibility issues and the limitations of datasets annotation preventing fair comparison across methods.}
    \label{fig:affordance_venn}
    \vspace{-5pt}
\end{figure}

Multiple methods considered grasping as functional to object picking~\cite{lenz2015deep,bohg2013data,fang2020graspnet,chu2018real,zhang2019roi, kumra2017robotic}. On the contrary, we base our definition of affordance on the functional interaction with an object~\cite{myers2015affordance,hassanin2021visual,osiurak2017affordance} (see Fig.~\ref{fig:whatwherehow}), considering grasping as part of an actions sequence to accomplish a higher-level task, e.g. pouring the content of a bottle implies grasping the bottle or opening a can implies grasping the tab.  
Given a high-level task the agent has to perform, we consider as a visual affordance the combination of the following three aspects: 
\begin{itemize}
    \item \textit{what}: the potential action on the most suitable objects in the image to accomplish the task;
    \item \textit{where}: the region where the agent will interact with the object through its hand; and
    \item \textit{how}: the most physically plausible hand pose to interact with the object.
\end{itemize} 
Conditioning the task with \textit{what}, \textit{where}, and \textit{how}, limits the potential actions, number of regions, and agent's hand poses to a close set relevant to complete the task. Our definition also enables the grouping and comparison of previous works, showing that an incomplete formulation of the affordance (maximum two aspects) was considered. 
None of the previous surveys~\cite{hassanin2021visual,chen2023survey,ardon2020affordances} discussed the limitations in the formulation of each task or provided a unifying view of visual affordance that enables an agent to interact with objects. Despite providing an overview of methods and datasets, previous surveys did not discuss the inconsistencies of training setups that undermine the reproducibility and fair comparison of affordance methods (see Fig.~\ref{fig:affordance_venn}).

In this paper, we unify the formulation for visual affordance prediction across its various tasks that were treated separately or appear disconnected in previous works and surveys. Through the lens of this unified formulation, we show the redefinition of the formulation for each task and systematically review related methods and datasets, highlighting similarities and limitations. We analyse the reproducibility issues of previous works and design the Affordance Sheet (inspired by Model Cards~\cite{mitchell2019model})\footnote{Project webpage: \url{https://apicis.github.io/aff-survey}} to overcome the reproducibility challenges while facilitating transparency of new visual affordance methods. 

\section{Problem formulation}
\label{sec:problem_formulation}

Let $x_v \in R^{F \times W \times H \times C}$ be the observed scene, where $F$ is the number of frames of an image sequence, $W$ is the image width, $H$ is the image height, $C$ is the number of channels ($C=3$ for an RGB input), and $v$ is the camera view index in a multi-camera setup. 
Let $\mathcal{T} = \{ t_m \, | \, t_{m-1} <  t_m < t_{m+1} \}^{M}_{m=1}$ be a task the agent needs to perform and represented by a sequence of steps $t_m$ expressed as text.
For example, a task could include the following steps: ``close the bottle'' and ``move the bottle to the shelf'', or only ``close the bottle''.
Let $\mathcal{E}$ be the set of hands (human or robotic) that can interact with the objects, and $e \in \mathcal{E}$ encode the characteristics of the agent's hand (size, number of fingers, and degrees of freedom) in a parametric model (e.g. MANO~\cite{romero2017embodied}). Let $\mathcal{O}$ be the set of objects relevant for the task (objects of interest). Objects can be localised using an intermediate model of object detection from the image $x_v$ and task $\mathcal{T}$. Each object $o \in \mathcal{O}$ can be represented as a bounding box $b \in \mathbb{R}^4$, indicating the position and size in $x_v$, an object class $\lambda$, and a confidence $c$: $ o = [b,\lambda, c]$. 
Let $\mathcal{A}_o$ be the set of potential actions that the agent performs on the object and each action $a \in \mathcal{A}_o$ can be expressed in text form. For example, for the task ``close the bottle'', the bottle cap affords the \textit{graspable} action.
Let $\mathcal{S}$ be the set of image regions on the object $o$ the agent can interact with to perform the action $a$. In general, to each action and object corresponds an interaction region $S_{o,a} \in \mathcal{S}$ on the objects of interest. $S$ can be represented as a probability map $[0, 1]^{W \times H}$ having zero values in the pixels belonging to the background and values greater than zero on the object pixels. To perform the action, the agent estimates how its hand interacts with the object, i.e. the pose of the hand $P$ on the interaction region of the object, also indicating how the fingers should close. For each object and action, the pose of the hand can be represented as a rotation-translation matrix $P = [R|T] \in SE(3)$, where $SE(3)$ is the special Euclidean group, $R \in SO(3)$ is a $3 \times 3$ rotation matrix in the special orthogonal group, and $T \in \mathbb{R}^3$ is the translation vector in the Euclidean space. With the pose, the hand can be rendered on the image plane to visualise the interaction with the object ($\tilde{x}_v \in R^{F \times W \times H \times C}$).

For a given task $\mathcal{T}$ and a visual input $x_v$, we define a visual affordance as a region $S$ that enables an agent with its hand $e$ to perform an action $a$ through a pose $P$ on a relevant object $o$. A visual affordance model is a function that maps the observed scene $x_v$, the task $\mathcal{T}$, and the hand $e$, into the objects of interest $o$, the potential action $a$, the regions of interaction $S$, and the pose of the end effector $P$:
\begin{equation}
    f(x_v, \mathcal{T}, e) \rightarrow \{ a, o, S, P \}.
\end{equation}

In this paper, we focus on methods for visual affordance prediction from an RGB image (visual input is $x_v = I; I \in R^{W \times H \times 3}$) and on the single hand case; we refer the reader to other works on affordance prediction from an RGB-D input~\cite{mousavian20196, jiang2022a4t,rosenberger2020object,yang2020human, fang2023anygrasp} and from multi-view inputs (including stereo)~\cite{pang2024stereo}. Removing the hand $e$, the task $\mathcal{T}$, or both, increases the number of possible solutions, making the problem too generic. An object can offer multiple actions for the same region, multiple regions can support the same action, and the region might not be realistic or feasible for specific agents (e.g. a robot with a 2-finger gripper). 

\section{Related works}
\label{sec:related_works}

\begin{table*}[t!]
    \centering
    \scriptsize
    \caption{Comparison of methods for object localisation.}
    \vspace{-5pt}
    \begin{tabular}{c c cc ccc c c c c}
    \toprule
    \textbf{Method} & \textbf{Source} & \multicolumn{2}{c}{\textbf{Backbone}} & \multicolumn{3}{c}{\textbf{Output}} & \textbf{GOR} & \textbf{AL} & \textbf{LLM} & \textbf{TAF}\\
    \cmidrule(lr){3-4} \cmidrule(lr){5-7}
    &  & Vision & Language & TC & BB & SEG & \\ 
    \midrule
    GGNN~\cite{sawatzky2019object} & GGNN~\cite{li2015gated} & RN-101~\cite{he2016deep} & - & \bbox & \bbox & \wbox & \bbox & \wbox & \wbox & \wbox\\ 
    TaskCLIP~\cite{chen2024taskclip} & CLIP~\cite{radford2021learning} & ViT-H~\cite{dosovitskiy2021image} & RoBERTa~\cite{liu2019roberta} & \wbox & \bbox & \wbox  & \wbox & \bbox & \bbox & \bbox \\
    VLTP~\cite{chen2024vltp} & SAM~\cite{kirillov2023segment} & ViT-H~\cite{dosovitskiy2021image} & SAM (encoder)~\cite{kirillov2023segment} & \wbox & \wbox & \bbox & \wbox & \wbox & \bbox & \bbox \\
    TOIST~\cite{li2022toist} & DETR~\cite{carion2020end} & RN-101~\cite{he2016deep} & RoBERTa~\cite{liu2019roberta} & \wbox & \bbox & \bbox & \wbox & \wbox & \wbox & \bbox \\ 
    CoTDet~\cite{tang2023cotdet} & DETR~\cite{carion2020end} & RN-101~\cite{he2016deep} & RoBERTa~\cite{liu2019roberta} & \wbox & \bbox & \bbox & \wbox & \wbox & \bbox & \bbox \\ 
    \bottomrule
    \addlinespace[\belowrulesep]
    \multicolumn{11}{l}{\parbox{0.8\linewidth}{\scriptsize{KEYS -- GOR:~graph-based objects relationship, AL:~vision-language alignment, LLM:~large language model, TAF:~attention-based fusion between task and vision features,  TC:~task classification, BB:~bounding box, SEG:~segmentation, RN:~ResNet, \bbox:~considered, \wbox:~not considered.}}}
    \end{tabular}
    \label{tab:relatedworks_taskdriven}
    \vspace{-8pt}
\end{table*}

Our formulation integrates the redefinitions related to affordance prediction given the task to accomplish and the RGB image. We decompose visual affordance prediction in the following subtasks and related components:
 \begin{enumerate}
     \item Localise the object of interest (\textit{object localisation}).
     \item Predict the actions for each localised object  (\textit{functional classification}).
     \item Predict the object regions that enable to perform the action (\textit{functional segmentation}).
     \item Estimate the agent's hand pose on the object, given the hand model and previous extracted information (\textit{hand pose estimation}).
     \item Render the hand on the RGB image (\textit{hand synthesis}).
 \end{enumerate}
 
Each component of our formulation instantiates one or more subtasks. 
For example, functional segmentation groups \textit{affordance segmentation} and \textit{affordance grounding}, as these subtasks have a similar problem formulation; \textit{grasping detection} can be considered as a special case of \textit{hand pose estimation}.
The components of our formulation provide the information about the target pose considering the desired result also from a visual point of view (rendered hand)

\subsection{Object localisation}

Given an image $I$ and a task $\mathcal{T}$, the model predicts a set of bounding boxes $\{ b_o\}^O_{o=1}$ with $b \in \mathbb{R}^4$, and a binary segmentation mask $\{ S_o\}^O_{o=1}$, with $S \in [0, 1]^{W \times H}$, 
\begin{equation}
    \{ b_o, S_o\}^O_{o=1} = f(I, \mathcal{T}) \:.
\end{equation}
The challenges of object localisation lie in fusing object appearance with context information, as some objects in the scene are not relevant for the task, while other objects are different but have similar functionality.

Methods either use a single architecture trained end-to-end~\cite{sawatzky2019object,li2022toist} or a combination of different models~\cite{chen2024taskclip,chen2024vltp,tang2023cotdet} (see Table~\ref{tab:relatedworks_taskdriven}). 
GGNN~\cite{sawatzky2019object} predicts the probability of each detected object being suitable for the task using Graph Neural Networks, where each node represents an object.
However, the assumption of a closed set of tasks and objects limits the generalization to unseen objects and unknown tasks. TOIST~\cite{li2022toist} overcomes the limitation of closed set of objects for each task through teacher-student training: the method replaces the object name in the student task sentence with an indefinite pronoun, replaces the pronoun token with the closest teacher token (nearest neighbour), and distils the teacher output.  
Other methods tackle the generalization to unseen objects and tasks, integrating Vision-Language Models (VLMs) or Large Language Models (LLMs)~\cite{chen2024taskclip,tang2023cotdet}. For example, CoTDet prompts an LLM to list the objects required to accomplish a task, the rationale that makes each object useful, and the object (textual) features. Cross-attention combines vision and textual tokens~\cite{tang2023cotdet} to predict the object bounding box. Alternatively, the cross-attention can combine vision and text tokens before the CLIP~\cite{radford2021learning} alignment, as in TaskCLIP~\cite{chen2024taskclip}, and a score function based on self-attention selects the objects that are more suitable for the task using the similarity matrix.

\subsection{Functional classification}

Functional classification, also referred to as affordance classification or affordance recognition, identifies \textit{what} are the potential actions (or affordance classes) $c$ that an agent can perform on an object from an input image $I$ given a task $\mathcal{T}$,
\begin{equation}
    \{ c_a\}^A_{a=1} = f(I, \mathcal{T}) \:.
\end{equation}

One of the main challenges of affordance classification is that without a defined task, one object has multiple affordances. For example, a cup on a table can suggest the action of picking or filling, but until a task is defined (e.g. `move the cup'), both affordances are plausible. 
Another challenge is that objects with similar appearances might afford different actions. For example, some models of trowel and turner might be similar in colour and shape, however the surface of a trowel is used to \textit{scoop}, while the surface of a turner to \textit{support}.

\begin{table}[t!]
    \centering
    \scriptsize
    \setlength\tabcolsep{2.5pt}
    \caption{Comparison of methods for functional classification. 
    Note that these methods use auxiliary tasks such as object detection or classification. 
    }
    \vspace{-5pt}
    \begin{tabular}{c c c c ccc}
    \toprule
    \textbf{Reference} & \textbf{Source} & \textbf{Backbone} & \textbf{Depth} & \textbf{DET} & \textbf{SEG} & \textbf{CLS} \\
    \midrule
    Nagarajan et al.~\cite{nagarajan2020ego} & - & ResNet~\cite{he2016deep} & \wbox & \wbox & \wbox & \wbox \\
    Sun et al.~\cite{sun2010learning} & PGM~\cite{jordan2004graphical} & - & \wbox & \wbox & \wbox & \bbox \\
    Zheng et al.~\cite{zheng2018high} & Faster R-CNN~\cite{ren2015faster} & VGG~\cite{simonyan2014very} & \wbox & \bbox & \wbox & \wbox \\
    Pieropan et al.~\cite{pieropan2013functional} &  SVM~\cite{cortes1995support} & - & \bbox & \bbox & \bbox & \wbox \\
    Kjellström et al.~\cite{Kjellstrom2011visual} & FCRF~\cite{sutton2004dynamic} & - & \wbox & \wbox & \wbox & \bbox \\
    \bottomrule
    \addlinespace[\belowrulesep]
    \multicolumn{7}{l}{\parbox{\columnwidth}{\scriptsize{KEYS -- Source:~source architecture, DET:~object detection, SEG:~object segmentation, CLS:~object classification, PGM:~probabilistic graphical model, SVM:~support vector machine, FCRF:~Factorial Conditional Random Field, \bbox:~considered, \wbox:~not considered.}}}
    \end{tabular}
    \label{tab:affordance-classification}
    \vspace{-8pt}
\end{table}

\begin{table*}[t!]
    \centering
    \scriptsize
    \caption{Comparison of visual affordance segmentation models~\cite{apicella2024segmenting}. We report the best-performing backbone for each model and do not consider additional parts of the pipelines, such as a separate object detector.}
    \vspace{-5pt}
    \begin{tabular}{c ccc c cccc cc ccc c}
    \toprule
    \textbf{Model} & \multicolumn{4}{c}{\textbf{Architecture}} & \multicolumn{4}{c}{\textbf{Attention}} & \multicolumn{2}{c}{\textbf{Affordance}} & \multicolumn{3}{c}{\textbf{Object}} & \textbf{CRF}\\ 
    \cmidrule(lr){2-5}\cmidrule(lr){6-9} \cmidrule(lr){10-11} \cmidrule(lr){12-14}
    & Source & Backbone & FPN & IF & Sp & Ch & Sa & Mc & CLA & ES & CLA & SEG & LOC & \\
    \midrule
    ADOSMNet~\cite{chen2023adosmnet} & PSPNet~\cite{zhao2017pyramid} & RN-101~\cite{he2016deep} & $\circ$ & $\circ$ & $\circ$ & $\circ$ & $\circ$ & $\circ$ & $\circ$ & $\circ$ & $\circ$ & $\circ$ & $\circ$ & $\circ$ \\ 
    CNN~\cite{nguyen2016detecting} & SegNet~\cite{badrinarayanan2015segnet} & VGG-16~\cite{simonyan2014very} & $\circ$ & $\circ$ & $\circ$ & $\circ$ & $\circ$ & $\circ$ & $\circ$ & $\circ$ & $\circ$ & $\circ$ & $\circ$ & $\circ$ \\ 
    RN50-F~\cite{hussain2020fpha} & Fast-FCN~\cite{wu2019fastfcn} & RN-50~\cite{he2016deep} & $\circ$ & $\circ$ & $\circ$ & $\circ$ & $\circ$ & $\circ$ & $\circ$ & $\circ$ & $\circ$ & $\circ$ & $\circ$ & $\circ$ \\ 
    BB-CNN~\cite{nguyen2017object} & DeepLab~\cite{chen2017deeplab} & VGG-16~\cite{simonyan2014very} & $\circ$ & $\circ$ & $\circ$ & $\circ$ & $\circ$ & $\circ$ & $\circ$ & $\circ$ & $\circ$ & $\circ$ & $\circ$ & $\bullet$ \\
    DeepLab~\cite{sawatzky2017weakly} & DeepLab~\cite{chen2017deeplab} & RN-101~\cite{he2016deep} & $\circ$ & $\circ$ & $\circ$ & $\circ$ & $\circ$ & $\circ$ & $\circ$ & $\circ$ & $\circ$ & $\circ$ & $\circ$ & $\bullet$ \\ 
    ACANet~\cite{apicella2023affordance} & UNet~\cite{ronneberger2015u} & RN-18~\cite{he2016deep} & $\circ$ & $\circ$ & $\circ$ & $\circ$ & $\circ$ & $\circ$ & $\circ$ & $\circ$ & $\circ$ & $\bullet$ & $\circ$ & $\circ$ \\ 
    AffordanceNet~\cite{do2018affordancenet} & Mask R-CNN~\cite{he2017mask} & VGG-16~\cite{simonyan2014very} & $\circ$ & $\circ$ & $\circ$ & $\circ$ & $\circ$ & $\circ$ & $\circ$ & $\circ$ & $\bullet$ & $\circ$ & $\bullet$ & $\circ$ \\ 
    4C-RPN-5C~\cite{minh2020learning} & AffordanceNet~\cite{do2018affordancenet} & SE-RNX-101~\cite{hu2018squeeze} & $\circ$ & $\circ$ & $\circ$ & $\circ$ & $\circ$ & $\circ$ & $\circ$ & $\circ$ & $\bullet$ & $\circ$ & $\bullet$ & $\circ$ \\ 
    B-Mask R-CNN~\cite{mur2023bayesian} & Mask R-CNN~\cite{morrison2019uncertainty,minh2020learning} & RNX-101~\cite{xie2017aggregated} & $\bullet$ & $\circ$ & $\circ$ & $\circ$ & $\circ$ & $\circ$ & $\circ$ & $\circ$ & $\bullet$ & $\circ$ & $\bullet$ & $\circ$ \\
    A-Mask R-CNN~\cite{caselles2021standard} & AffordanceNet~\cite{do2018affordancenet} & RN-50~\cite{he2016deep} & $\bullet$ & $\circ$ & $\circ$  & $\circ$ & $\circ$ & $\circ$ & $\circ$ & $\circ$ & $\bullet$ & $\circ$ & $\bullet$ & $\circ$ \\ 
    \midrule
    GSE~\cite{zhang2022multi} & HRNet~\cite{sun2019deep,zhang2017deep} & RNS-101~\cite{zhang2022resnest} & $\circ$ & $\bullet$ & $\circ$ & $\bullet$ & $\circ$ & $\circ$ & $\circ$ & $\circ$ & $\circ$ & $\circ$ & $\circ$ & $\circ$ \\ 
    DRNAtt~\cite{gu2021visual} & DANet~\cite{fu2019dual}  & DRN~\cite{yu2017dilated} & $\circ$ & $\circ$ & $\bullet$ & $\bullet$ & $\circ$ & $\circ$ & $\circ$ & $\circ$ & $\circ$ & $\circ$ & $\circ$ & $\circ$ \\
    SEANet~\cite{yin2022new} & DFF~\cite{hu2019dynamic} & RN-50~\cite{he2016deep} & $\circ$ & $\bullet$ & $\bullet$ & $\bullet$ & $\circ$ & $\circ$ & $\circ$ & $\bullet$ & $\circ$ & $\circ$ & $\circ$ & $\circ$ \\ 
    BPN~\cite{yin2022object} & AffordanceNet~\cite{do2018affordancenet} & RN-50~\cite{he2016deep} & $\bullet$ & $\circ$ & $\bullet$ & $\bullet$ & $\circ$ & $\circ$ & $\circ$ & $\bullet$ & $\bullet$ & $\circ$ & $\bullet$ & $\circ$ \\         
    RANet~\cite{zhao2020object} & EncNet~\cite{zhang2018context} & RN-50~\cite{he2016deep} & $\circ$ & $\circ$ & $\circ$ & $\bullet$ & $\circ$ & $\circ$ & $\bullet$ & $\circ$ & $\bullet$ & $\circ$ & $\circ$ & $\circ$ \\ 
    STRAP~\cite{cui2023strap} & SINN~\cite{nauata2019structured} & RN-50~\cite{he2016deep} & $\circ$ & $\bullet$ & $\circ$ & $\circ$ & $\bullet$ & $\circ$ & $\bullet$ & $\circ$ & $\circ$ & $\circ$ & $\circ$ & $\bullet$ \\
    M2F-Aff~\cite{apicella2024segmenting} & Mask2Former~\cite{cheng2022masked} & RN-50~\cite{he2016deep} & $\bullet$ & $\circ$ & $\circ$ & $\circ$ & $\bullet$ & $\bullet$ & $\bullet$ & $\circ$ & $\circ$ & $\circ$ & $\circ$ & $\circ$ \\ 
    \bottomrule
    \addlinespace[\belowrulesep]
    \multicolumn{15}{l}{\parbox{0.9\linewidth}{\scriptsize{KEYS -- Source:~reference architecture, Backbone:~visual encoder, Sp:~spatial attention, Ch:~channel attention, Mc: masked cross-attention, Sa: self-attention, CLA:~classification, ES:~edge segmentation, SEG:~segmentation, LOC:~localisation, RN:~ResNet, RNX:~ResNeXt, RNS:~ResNeSt, SE-RNX:~squeeze and excite ResNeXt, DRN:~Dilated Residual Network, CRF:~conditioned random fields, IF:~intermediate feature maps fusion; $\bullet$:~considered, $\circ$:~not considered}.}}
    \end{tabular}
    \label{tab:aff_seg_related_works}
    \vspace{-8pt}
\end{table*}

Methods for affordance classification learn actions that can be performed with objects in the scene either from human demonstration~\cite{pieropan2013functional,nagarajan2020ego,Kjellstrom2011visual}, or from images of the environment~\cite{sun2010learning,zheng2018high}. We summarise the characteristics of these methods in Table~\ref{tab:affordance-classification}. Nagarajan et al.~\cite{nagarajan2020ego} trained an affordance classifier to predict all the potential actions that a person can perform in an environment (e.g., a kitchen sink). Sun et al.~\cite{sun2010learning} used Probabilistic Graphical models to relate object affordances with appearance. Images are processed with dimensionality reduction, limiting the scalability of the method to high resolution images, and increasing the complexity of the graph structure adding affordance categories. The combination of affordance classification with auxiliary tasks such as detection and segmentation allows to focus only on regions of interest in the image and to group objects based on the actions they are used for (functionality), instead of their appearance~\cite{pieropan2013functional,zheng2018high}. 
By training methods on data of people using objects or with the agent exploring the environment, previous works~\cite{pieropan2013functional,nagarajan2020ego,sun2010learning,Kjellstrom2011visual} implicitly considered as a task the functional use of the object.
However, these methods do not consider the physical interaction between the agent and the object, as the action is not associated with an interaction region in the image~\cite{sun2010learning,zheng2018high,Kjellstrom2011visual}. This results in the agent having multiple options (ambiguity) on \textit{how} and \textit{where} to perform the interaction. For example, even a simple instruction like ``move the cup'' can be performed in multiple ways, such as grasping the cup by the body or by the rim.

\subsection{Functional segmentation} 

The segmentation of functional regions on objects in the image identifies \textit{where} the agent needs to perform the interaction with the object. This is approached in two ways (see Table~\ref{tab:aff_seg_related_works}). 
\textit{Affordance detection and segmentation} detects the objects of interest in the image and separates the functional regions.  \textit{Affordance grounding} identifies on the object the region that should be used to perform the action defined in the task. 

\noindent{\bf Affordance detection and segmentation.} 
Given an image $I$, the model predicts bounding boxes $\{ b_o\}^O_{o=1}$ and segmentation masks of $A$ functional regions $\{ S_o\}^O_{o=1}$ for objects of interest, 
\begin{equation}
    \{ b_o, S_o \}^O_{o=1} = f(I, \mathcal{T}) \:.
\end{equation}
The segmentation mask $S_o$ can be also formulated as the combination of the actions $\{ c_a \}^A_{a=1}$ with a probability map $\{ S_{o,a}\}$ where $S \in [0, 1]^{W \times H}$ indicates the region where an action takes place for each object~\cite{apicella2023affordance, cui2023strap}.
Affordance detection and segmentation methods assume that the objects of interest are the ones annotated in the dataset, that the task $\mathcal{T}$ is to use the object to fulfil the purpose it was designed for ~\cite{nguyen2017object,myers2015affordance,do2018affordancenet,nguyen2016detecting}, and that different parts of the objects are associated with a functionality to accomplish the task. For example, in a knife the handle is designed to be grasped while the blade is used for cutting another object. These methods localise the object that affords ``cut'' (detection) and segment the blade that affords the action ``cut'' (segmentation). 

Previous methods~\cite{zhao2017pyramid,badrinarayanan2015segnet,wu2019fastfcn,chen2017deeplab,ronneberger2015u,he2017mask,fu2019dual} adapt \textit{semantic} and \textit{instance segmentation} architectures to predict affordance regions on the objects. For example, A-Mask R-CNN~\cite{caselles2021standard} and AffordanceNet~\cite{do2018affordancenet} modify an instance segmentation model (Mask R-CNN~\cite{he2017mask}) to predict affordance masks instead of object masks for each localised object. Starting from the design of AffordanceNet, BPN~\cite{yin2022object}, and 4C-RPN-5C~\cite{minh2020learning} combine the region of interest with the feature maps at different resolutions and predict the overlapping of bounding boxes and boundaries of affordance regions. The object detection branch localises regions of interest in the image, but inaccurate or wrong predictions can consequently result in segmenting affordance regions outside of the actual objects. When edges are blurred or not clearly defined (e.g. occlusions or transparent objects), BPN fails to predict precise affordance contours despite its additional edge segmentation component. On the contrary, semantic segmentation models~\cite{nguyen2016detecting,hussain2020fpha,apicella2023affordance, zhang2022multi,gu2021visual, yin2022new, zhao2020object} avoid the dependence from an object detector and assign each pixel of the image to an affordance class (per-pixel affordance segmentation). When objects are occluded or boundaries are not clearly defined, methods such as CNN~\cite{nguyen2016detecting}, RN50-F~\cite{hussain2020fpha}, and ACANet~\cite{apicella2023affordance} can classify affordance pixels outside the object region.

\textit{Attention mechanisms}~\cite{zhang2022multi,gu2021visual,yin2022new,yin2022object,zhao2020object} are an alternative way to consider only relevant information in the image by weighing image feature maps. 
For example, GSE~\cite{zhang2022multi}, DRNAtt~\cite{gu2021visual}, SEANet~\cite{yin2022new}, and BPN~\cite{yin2022object}, learn the channels weight or the similarity between positions in the feature map without direct supervision. 
For computational reasons, both DRNAtt and GSE process feature maps at low-resolutions where important details for affordance segmentation (e.g. edges) are degraded for objects not in foreground. 
In RANet~\cite{zhao2020object}, the attention weights are learned with the supervision of object classes. However, in case of occlusions, mistakes in the attention weights cause mismatch between the predicted object classes and the segmented affordances.

Most of previous methods~\cite{nguyen2016detecting,hussain2020fpha,apicella2023affordance,gu2021visual, yin2022new, zhao2020object, zhang2022multi} performed the classification of the affordances and the segmentation of regions jointly. 
However, the two subtasks can be decoupled 
assigning an affordance class to each segmentation mask~\cite{cui2023strap}. For example, STRAP~\cite{cui2023strap} learns the affordance classification and segmentation in separate branches. The model learns to segment affordance masks with weak supervision from a point annotation of each region~\cite{tang2018regularized} and by using Conditional Random Fields to process the pixel position and colour. However, this approach can lead to inaccurate segmentations when the object colour is not clearly distinguished from the background~\cite{sawatzky2017weakly,nguyen2016detecting}. STRAP also uses self-attention to process low-resolution image feature maps, 
losing details about the object in the image when the object scale is small. To increase the resolution of processed feature maps, M2F-AFF~\cite{apicella2024segmenting} adapted Mask2Former~\cite{cheng2022masked}  
that combines the image features with learnable latent vectors, while ignoring the pixel positions outside the object region (background) through masked cross-attention. 

Note that affordance segmentation is tackled independently from the agent's hand characteristics, even if the number of fingers or the degrees of freedom influence the contact regions on the object. Nevertheless, affordance regions can be used by an agent such as a robot to perform actions~\cite{nguyen2017object, do2018affordancenet, yin2022object}.

\noindent{\bf Affordance grounding.} 
Given an image $I$ and a task $\mathcal{T}$, the model predicts the probability map $\{ S_o\}^O_{o=1}$ identifying the region that the robot can use to interact with the object,
\begin{equation}
    \{ S_o \}^O_{o=1} = f(I, \mathcal{T}) \:.
\end{equation}
$\mathcal{T}$ can be expressed through natural language~\cite{qian2024affordancellm, qu2024knowledge}, an affordance category~\cite{luo2022learning,nagarajan2019grounded,tong2024oval,li2024one}, a point in 2D~\cite{qian2023understanding,cuttano2024does}, or another image of the object of interest~\cite{li2023locate,hadjivelichkov2023one,demo2vec2018cvpr}.
With this formulation, affordance grounding and one-shot methods using prior information can be grouped together.  

\begin{table}[t!]
    \centering
    \scriptsize
    \setlength\tabcolsep{1.5pt}
    \caption{Comparison of affordance grounding methods.}
    \vspace{-5pt}
    \begin{tabular}{c cccc c c c cc}
    \toprule
    \textbf{Method} & \multicolumn{4}{c}{\textbf{Prior}} & \textbf{Vid-img} & \textbf{Exo-ego} & \textbf{CAM} & \multicolumn{2}{c}{\textbf{Supervision}}\\
    \cmidrule(lr){2-5} \cmidrule(lr){9-10} 
    & 2D-P & IMG & CLS & Task & & & & strong & weak \\
    \midrule
    3DOI~\cite{qian2023understanding} & \bbox & \wbox & \wbox & \wbox & \wbox & \wbox & \wbox & \bbox & \wbox \\
    CALNet~\cite{luo2023leverage} & \wbox & \bbox & \wbox & \wbox & \wbox & \bbox & \wbox & \bbox & \wbox \\
    LOCATE~\cite{li2023locate} & \wbox & \bbox & \wbox & \wbox & \wbox & \bbox & \bbox & \wbox & \bbox\\
    AffCorrs~\cite{hadjivelichkov2023one} & \wbox & \bbox & \wbox & \wbox & \wbox & \wbox & \wbox & \wbox & \wbox\\
    Demo2Vec~\cite{demo2vec2018cvpr} & \wbox & \bbox & \wbox & \wbox & \bbox & \wbox & \wbox & \bbox & \wbox \\
    Hotspots~\cite{nagarajan2019grounded} & \wbox & \wbox & \bbox & \wbox & \bbox & \wbox  & \bbox & \wbox & \bbox\\
    Cross-View-AG~\cite{luo2022learning, luo2024grounded} & \wbox & \wbox & \bbox & \wbox & \wbox & \bbox  & \bbox & \wbox & \bbox \\
    OVAL-Prompt~\cite{tong2024oval} & \wbox & \wbox & \bbox & \wbox & \wbox & \wbox  & \wbox & \wbox & \wbox\\
    AffordanceCLIP~\cite{cuttano2024does} & \wbox & \wbox & \bbox & \wbox & \wbox & \wbox  & \wbox & \bbox & \wbox\\
    OOAL~\cite{li2024one} & \wbox & \wbox & \bbox & \wbox & \wbox & \wbox  & \wbox & \bbox & \wbox \\
    KBAG-Net~\cite{qu2024knowledge} & \wbox & \wbox & \wbox & \bbox & \wbox & \wbox & \wbox & \bbox & \wbox \\
    AffordanceLLM~\cite{qian2024affordancellm} & \wbox & \wbox & \bbox & \bbox & \wbox & \wbox & \wbox & \bbox & \wbox \\
    \bottomrule
    \addlinespace[\belowrulesep]
    \multicolumn{10}{l}{\parbox{\columnwidth}{\scriptsize{KEYS -- Vid-img:~transfer from video to image, Exo-ego:~transfer from exocentric to egocentric view, CAM:~Class Activation Maps, 2D-P:~point in image, IMG:~a support image/region, CLS:~action class, \bbox:~considered, \wbox:~not considered.}}}
    \end{tabular}
    \label{tab:affordance-grounding}
    \vspace{-8pt}
\end{table}

We summarise the characteristics of affordance grounding methods in Table~\ref{tab:affordance-grounding}. 
Formulating the task as an additional input to the model enables affordance grounding methods to tackle generalization to object categories while avoiding an explicit object detection phase.
Methods for explainability (Class Activation Maps~\cite{zhou2016learning}) highlight the region in the image that corresponds to the action~\cite{luo2022learning,nagarajan2019grounded,li2023locate,luo2024grounded}. However, these regions are not bounded by object contours, limiting the application of these methods to unoccluded object settings. One-shot-based methods use an image as a prior to select objects of interest~\cite{luo2021one,zhai2022one}, or segment affordance regions~\cite{hadjivelichkov2023one}, based on the similarity between the input images and the prior (query image). However, the support image is assumed to be similar to the query images, thus implying that the object category in the scene should be known in advance.

To cope with the limited amount of training images, methods adapt pre-trained models~\cite{qian2024affordancellm,tong2024oval,li2024one,cuttano2024does}, using knowledge transfer from video to image~\cite{nagarajan2019grounded,demo2vec2018cvpr} or from exocentric to egocentric views of the object~\cite{luo2022learning,li2023locate,luo2023leverage}. In particular, multimodal models help generalising to unknown object categories or unknown actions (open vocabulary). For example, AffordanceCLIP adapts CLIP~\cite{radford2021learning} with a learnable feature pyramid network to predict the affordance probability map~\cite{cuttano2024does}. A contrastive loss encourages the alignment between pixel-level embeddings within the annotated mask of the object and language features. AffordanceLLM~\cite{qian2024affordancellm} processes vision and language information to predict affordance segmentation tokens. The LLM generates text tokens encoding the object part used to perform the task and a mask token that is combined with the visual tokens using a transformer decoder to predict the affordance map. KBAG-Net~\cite{qu2024knowledge} fuses language features extracted using BERT~\cite{devlin2019bert} with low and high resolution features from a visual backbone~\cite{he2016deep}. A convolutional decoder processes fused features to predict the affordance map. 

Few of the methods~\cite{luo2022learning,li2023locate, luo2023leverage} for affordance grounding focused on learning object affordances by building correspondences from the exocentric view of an object (human using the object) to the egocentric view (object only). Both LOCATE~\cite{li2023locate} and Cross-View-AG~\cite{luo2022learning} during training combined a loss to learn the affordance category with losses to preserve the similarity between the feature maps of the exocentric and egocentric views.  
CALNet~\cite{luo2023leverage} models the correspondence between contact regions in the exocentric and egocentric views, concatenating the human keypoint features extracted from the exocentric view with the visual features of the egocentric perspective.
Instead of learning directly from images, methods like Demo2Vec~\cite{demo2vec2018cvpr} and Hotspots~\cite{nagarajan2019grounded} learn to transfer the affordance from videos of humans interacting with objects in household environments, e.g. oven, fridge, washing machine, to the images containing only the objects. Although these methods~\cite{luo2022learning,nagarajan2019grounded,li2023locate,demo2vec2018cvpr,luo2023leverage} can learn the object affordances from examples showing humans that perform actions, the egocentric views are composed by the object on a background without occlusions or clutter, limiting the generalisation to in-the-wild images.

Despite generalising to different actions or task formulations, affordance grounding methods output confidence maps that are not bounded by object edges. The confidence maps could also overlap with other objects in case of clutter or with a human hand if the object is hand-held. Using a coarse confidence map when interacting with an object can lead an agent to misplace its hand, thus undermining the success of the interaction or harming the human.

\subsection{Hand pose estimation and synthesis}

To perceive the visual affordance, the agent predicts also \textit{how} to perform the interaction with the object, i.e. the pose of the agent hand. Previous works~\cite{lenz2015deep,corona2020ganhand,chu2018real, zhang2019roi, kumra2017robotic, lundell2021multi, redmond2015real, asif2018graspnet, ainetter2021end} related the problem mostly to grasping rather than to visual affordances, and redefined the problem based on the hand: \textit{grasping detection} for two-finger grippers~\cite{lenz2015deep,chu2018real,zhang2019roi,kumra2017robotic,asif2018graspnet}, \textit{hand-object pose estimation} for human hand~\cite{ye2023affordance,corona2020ganhand} (e.g. MANO model~\cite{romero2017embodied}), \textit{multi-finger grasping} for three fingers Barrett hand~\cite{lundell2021multi}. 
Given the hand model, the image of the object, and the task, the model predicts the pose $P$ of the hand on the object,
\begin{equation}
    \{ P_o \}^O_{o=1} = f(I, \mathcal{T}, e) \:.
\end{equation}
Predicting the pose of the agent's hand, however, is challenging because the hand is not observed in the image, and therefore only the visual features of the object can be used. 

\noindent{\bf Grasping detection.} 
Assuming a two-finger gripper, the pose estimation is reformulated as prediction of grasping points directly on the image, encoding the parameters of the gripper as an oriented rectangle~\cite{lenz2015deep}. 
In particular, 1 DoF encodes the rotation with respect to the horizontal axis, 2 DoF encode the translation of the gripper centre (horizontal and vertical), and 2 DoF encode the geometry (opening width and fingers height). The underlying assumption is the availability of a depth map to obtain the full 7 DoF representation of the gripper in 3D (translation, rotation, and opening width).
Given an RGB-D image $I \in \mathbb{R}^{W \times H \times 4}$, the model predicts a set of $G$ oriented rectangles $\{r_g\}^G_{g=1}$ with $r \in \mathbb{R}^5$ consisting of the rectangle centre coordinates, the rectangle size (width and height), and the orientation. 
Predicting the pose of a two-finger gripper on 
an object is challenging because each object has multiple grasping points, but only a subset of grasping poses leads to successful grasping. Moreover, when estimating the grasping points from a single view, only a side of the object is visible, limiting the number of feasible grasping points.

\begin{table}[t!]
    \centering
    \scriptsize
    \setlength\tabcolsep{1pt}
    \caption{Comparison of grasping detection models.
    }
    \vspace{-5pt}
    \begin{tabular}{c c c c cc cccc}
    \toprule
    \textbf{Method} & \textbf{Backbone} &  \textbf{D} & \textbf{2stages} & \multicolumn{2}{c}{\textbf{Modality fusion}} & \multicolumn{4}{c}{\textbf{Auxiliary tasks}} \\
    \cmidrule(lr){5-6} \cmidrule(lr){7-10}
    & & & & EAR & MID & GLIKE & GSEG & DET & SEG\\
    \midrule
    MultiGrasp~\cite{redmond2015real} & AXN~\cite{krizhevsky2012imagenet} & $\bullet$  & $\circ$ & $\circ$ & $\circ$ & $\bullet$ & $\circ$ & $\circ$ & $\circ$ \\
    Kumra et al.~\cite{kumra2017robotic} & RN-50~\cite{he2016deep} & $\bullet$ & $\circ$ & $\circ$ & $\bullet$ & $\circ$ & $\circ$ & $\circ$ & $\circ$ \\
    GraspNet~\cite{asif2018graspnet} & - & $\bullet$ & $\circ$ & $\circ$ & $\circ$ & $\circ$ & $\bullet$ & $\circ$ & $\circ$ \\
    Ainetter et al.~\cite{ainetter2021end} & RN-101~\cite{he2016deep} & $\circ$ & $\bullet$ & $\circ$ & $\circ$ & $\circ$ & $\circ$ & $\circ$ & $\bullet$ \\
    Lenz et al.~\cite{lenz2015deep} &  - & $\bullet$ & $\bullet$ & $\bullet$ & $\circ$ & $\circ$ & $\circ$ & $\circ$ & $\circ$ \\
    Chu et al.~\cite{chu2018real} & RN-50~\cite{he2016deep} & $\bullet$ & $\bullet$ & $\circ$ & $\circ$ & $\circ$ & $\circ$ & $\circ$ & $\circ$ \\
    ROI-GD~\cite{zhang2019roi} & RN-101~\cite{he2016deep} & $\bullet$ & $\bullet$ & $\circ$ & $\circ$ & $\circ$ & $\circ$ & $\bullet$ & $\circ$ \\
    \bottomrule
    \addlinespace[\belowrulesep]
    \multicolumn{10}{l}{\parbox{\columnwidth}{\scriptsize{KEYS -- D:~depth, EAR:~early fusion, MID:~middle fusion, GLIKE:~grasp likelihood, GSEG:~grasping segmentation, DET:~object detection, SEG:~object segmentation, AXN:~AlexNet, RN:~ResNet, $\bullet$:~considered, $\circ$:~not considered.}}}
    \end{tabular}
    \label{tab:relatedworks_grasping}
    \vspace{-8pt}
\end{table}

Table~\ref{tab:relatedworks_grasping} summarises methods for grasping detection. 
Most of the methods~\cite{lenz2015deep,chu2018real,zhang2019roi,kumra2017robotic,asif2018graspnet} use RGB-D images to predict grasping rectangles, as the depth information provides geometric cues. Visual information is fused in different ways: in the first layers of the model~\cite{lenz2015deep,chu2018real,zhang2019roi,asif2018graspnet} (early fusion), or using a separate backbone to process RGB and depth before fusion (middle fusion)~\cite{kumra2017robotic}. However, there are no results showing that a fusion mechanism is more effective than the others.
The feature extraction is performed mainly by convolutional networks like ResNet~\cite{chu2018real,zhang2019roi,kumra2017robotic,ainetter2021end} or AlexNet~\cite{redmond2015real} pre-trained on ImageNet~\cite{krizhevsky2012imagenet}, to transfer the features learned on large scale datasets.
Methods can be categorised into single-stage and two-stage: single-stage methods~\cite{kumra2017robotic,redmond2015real,asif2018graspnet} predict the final oriented rectangles from the image, either directly regressing the rectangle~\cite{kumra2017robotic,redmond2015real} or considering the rectangle as a by-product of object segmentation~\cite{asif2018graspnet}; two-stage methods~\cite{lenz2015deep,chu2018real,zhang2019roi,ainetter2021end} first predict grasping candidates (coarse estimation) and then refine the predictions (fine estimation).
The majority of two-stage methods adapt works for object detection (e.g. Faster R-CNN~\cite{ren2015faster}) to grasping detection in different ways: separating the learning of the object location and grasp locations~\cite{zhang2019roi}; separating the learning of the quantized angle from the learning of the centre, width and height of the grasping rectangle~\cite{chu2018real}; separating the coarse prediction of grasping rectangles from the refinement based on the object segmentation~\cite{ainetter2021end}.   
Auxiliary tasks, such as object detection~\cite{chu2018real} and segmentation~\cite{ainetter2021end} constrain the prediction of the grasping rectangle to the object, reducing mistakes in cluttered scenes or when the object is not in foreground and completely visible. Other auxiliary tasks are: the likelihood of an image patch (non-overlapping piece of the image) containing a grasp~\cite{redmond2015real} limiting the prediction of grasping rectangles to some parts of the image; and the grasping region segmentation~\cite{asif2018graspnet} constraining the grasping rectangle to graspable region of the object, e.g. the handle of a spoon.

Grasping detection formulation considers only the interaction of picking, resulting in non-functional solutions as the agent can grasp the object at any surface location. For example, the rim of a cup filled with liquid (suggesting the affordance of pouring the content) might be selected as a potential grasping point without considering that the liquid might be spilled or damage the robotic hand. Most of the methods for grasping detection~\cite{lenz2015deep,chu2018real,zhang2019roi,kumra2017robotic,redmond2015real,depierre2018jacquard} assume that objects are observed on a tabletop or on the floor (top-down camera view). Hence, models fail to generalise to scenarios with different camera view-points or with occlusions.

\noindent{\bf Multi-finger pose estimation and interaction synthesis.} Previous works~\cite{ye2023affordance,corona2020ganhand,lundell2021multi} considered as visual affordance the pose of an agent's hand on objects in the scene. 
Given an image $I$, the model predicts the 6D pose of the hand on the object $\{[R|T]_o\}^O_{o=1}$ with $[R|T]$ representing pose of the hand  and renders an image of the hand, $\tilde{I} \in \mathbb{R}^{W \times H \times 3}$ (interaction synthesis), showing \textit{how} and \textit{where} the hand interacts with the object, not \textit{what} action is performed.

\begin{table}[t!]
    \centering
    \scriptsize
    \setlength\tabcolsep{2pt}
    \caption{Comparison of multi-finger pose estimation and interaction synthesis methods.}
    \vspace{-5pt}
    \begin{tabular}{c c cc cc}
    \toprule
    \textbf{Method} & \textbf{Obj. pose} & \multicolumn{2}{c}{\textbf{Grasp}} & \multicolumn{2}{c}{\textbf{Learning}} \\
    \cmidrule(lr){3-4} \cmidrule(lr){5-6}
    & & CLS & LOC & ADV & DIFF \\
    \midrule
    Multi-FinGAN~\cite{lundell2021multi} & $\bullet$ & $\bullet$ & $\circ$ & $\bullet$ & $\circ$ \\
    GanHand~\cite{corona2020ganhand} & $\bullet$ & $\bullet$ & $\circ$ & $\bullet$ & $\circ$\\
    AffordanceDiffusion~\cite{ye2023affordance} & $\circ$ & $\circ$ & $\bullet$ & $\circ$ & $\bullet$\\
    \bottomrule
    \addlinespace[\belowrulesep]
    \multicolumn{6}{l}{\parbox{0.7\columnwidth}{\scriptsize{KEYS -- CLS:~category, LOC:~location, ADV:~adversarial based, DIFF:~diffusion based, $\bullet$:~considered, $\circ$:~not considered}.}}
    \end{tabular}
    \label{tab:pose-estimation}
    \vspace{-10pt}
\end{table}

Table~\ref{tab:pose-estimation} compares the characteristics of methods for multi-finger pose estimation and interaction synthesis. Methods are based on a coarse-to-fine approach locating first where the hand will interact with the object and then refining the pose using generative adversarial networks~\cite{corona2020ganhand, lundell2021multi} or diffusion models~\cite{ye2023affordance}. GanHand~\cite{corona2020ganhand} estimates objects’ shapes and locations using an object 6D pose estimator or a reconstruction network. GanHand localises the object projecting its shape in the image plane and predicts the grasp type, i.e. the type of interaction between hand and object. The network predicts the coarse pose of the hand from the grasp type and the visual features, and refines the hand parameters to obtain the final shapes and poses (i.e. MANO model~\cite{romero2022embodied}), learned by minimising an adversarial loss with a discriminator. 
Multi-FinGAN~\cite{lundell2021multi} adapts GanHand architecture to perform the pose estimation of the Barrett end-effector on the object in the image. Contrary to GanHand, Multi-FinGAN uses the object reconstruction only to refine the coarse pose of the end-effector. As a consequence, the method underperforms if multiple objects are present in the scene.
AffordanceDiffusion~\cite{ye2023affordance} is a cascade of two diffusion models to generate the image of the hand interacting with the object in the image. The diffusion process uses a prior (forearm mask) composed by a circle representing the hand and a rectangle representing the forearm. For every diffusion step, the first model predicts the denoised forearm mask from the features of the forearm mask obtained in the previous step, the object image, and the forearm mask projected on the object image. The second model combines the layout mask (prior) with the object image to synthesise the interaction.

Methods for multi-finger pose estimation and synthesis focus on a generic grasping interactions, without taking into account the task that the agent performs and the affordances that the object supports. This fact can result in estimating wrong poses not aligned with the task. 
 
\section{Datasets: review and limitations}
\label{sec:review_datasets}

In this section, we compare the characteristics of image-based datasets for visual affordance prediction and discuss their similarities and limitations (see Table~\ref{tab:review_datasets}), contrary to previous surveys~\cite{hassanin2021visual,chen2023survey}. Our comparison considers elements such as: the type of environment (indoor or outdoor); the camera viewpoint (third person or first person); the objects of interest (quantity, diversity depending on the group such as tools or containers, physical properties such as transparency); the type of images (real, simulated, mixed-reality); the presence of occlusions, due to clutter, or hand manipulating the object; and the annotations of affordances (quantity, accuracy, procedure, and expertise of the annotators). 
These datasets are usually split into two non-overlapping sets: one to train models (training set) and another to evaluate their performance (testing set). During training, biases in the images, ambiguities or inaccuracies in the annotations are transferred to the models.

\begin{table}[t!]
    \scriptsize
    \setlength\tabcolsep{1.5pt}
    \centering
    \caption{Characteristics of datasets for visual affordance prediction grouped by task.}
    \vspace{-5pt}
    \begin{tabular}{c c r cc c c c c}
         \toprule
         \textbf{Task} & \textbf{Dataset} & \textbf{\# Images} & \textbf{OBJ} & \textbf{AFF} & \textbf{Real} & \textbf{Tran.} & \textbf{3PV} & \textbf{HOc} \\
         \midrule
         \multirow{2}{*}{\textit{OBJD}}
         & Rio~\cite{qu2023rio} & 40,214 & - & - & \bbox & \wbox & \bbox & \wbox \\
         & COCO-Task~\cite{sawatzky2019object} & 39,724 & 49 & - & \bbox & \wbox & \bbox & \wbox \\
         \midrule
         \multirow{4}{*}{\textit{AFFC}}
         & Pieropan et al.~\cite{pieropan2013functional} & $\sim$40,000 & 4 & 4 & \bbox & \wbox & \bbox & \wbox \\ 
         & Zheng et al.~\cite{zheng2018high} & 740 & 8 & 3 & \bbox & \wbox & \bbox & \wbox \\ 
         & Sun et al.~\cite{sun2010learning} & ~1400 & 7 & 6 & \bbox & \wbox & \bbox & \wbox \\ 
         & Kjellström et al.~\cite{Kjellstrom2011visual} & ~11,500 & 6 & 3 & \bbox & \wbox & \bbox & \raisebox{1pt}{\scalebox{0.5}{\LEFTcircle}} \\
         \midrule
         \multirow{2}{*}{\textit{AFFG}}
         & OPRA~\cite{demo2vec2018cvpr} & - & - & 7 & \bbox & \wbox & \bbox & \wbox \\
         & AGD20K~\cite{luo2022learning} & 23,816 & 47 & 36 & \bbox & \raisebox{1pt}{\scalebox{0.5}{\LEFTcircle}} & \bbox & \bbox \\
         \midrule
         \multirow{10}{*}{\textit{AFFDS}}
         & AFF-Synth~\cite{christensen2022learning} & 30,245 & 21 & 7 & \wbox & \wbox & \bbox & \wbox \\
         & UMD-Synth~\cite{chu2019learning} & 37,200 & 17 & 7 & \wbox & \wbox & \bbox & \wbox \\
         & Multi-View~\cite{khalifa2022towards} & 47,210 & 37 & 15 & \bbox & \wbox & \bbox & \wbox \\
         & HANDAL~\cite{guo2023handal} & 308,000 & 17 & 1 & \bbox & \wbox & \bbox & \raisebox{1pt}{\scalebox{0.5}{\LEFTcircle}}  \\
         & TRANS-AFF~\cite{jiang2022a4t} & 1,346 & 3 & 3 & \bbox & \bbox & \bbox & \wbox \\
         & UMD~\cite{myers2015affordance} & 28,843 & 17 & 7 & \bbox & \wbox & \bbox & \wbox \\ 
         & IIT-AFF~\cite{nguyen2017object} & 8,835 & 10 & 9 & \bbox & \raisebox{1pt}{\scalebox{0.5}{\LEFTcircle}} & \raisebox{1pt}{\scalebox{0.5}{\LEFTcircle}} & \raisebox{1pt}{\scalebox{0.5}{\LEFTcircle}} \\
         & CAD120-AFF~\cite{sawatzky2017weakly} & 3,090 & 11 & 6 & \bbox & \wbox & \bbox & \raisebox{1pt}{\scalebox{0.5}{\LEFTcircle}} \\ 
         & FPHA-AFF~\cite{hussain2020fpha} & 4,300 & 14 & 8 & \bbox & \raisebox{1pt}{\scalebox{0.5}{\LEFTcircle}} & \wbox & \bbox \\
         & EPIC-AFF~\cite{mur2023multi} & 38,876 & 304 & 43 & \bbox & \raisebox{1pt}{\scalebox{0.5}{\LEFTcircle}} & \wbox & \bbox \\
         & CHOC-AFF~\cite{apicella2023affordance} & 138,240 & 3 & 3 & \raisebox{1pt}{\scalebox{0.5}{\LEFTcircle}} & \raisebox{1pt}{\scalebox{0.5}{\LEFTcircle}} & \bbox & \raisebox{1pt}{\scalebox{0.5}{\LEFTcircle}} \\
         \midrule
         \multirow{4}{*}{\textit{GDET}}
         & Cornell grasping~\cite{jiang2011efficient} & 1,035 & - & 1 & \bbox & \wbox & \bbox & \wbox \\ 
         & GraspSeg~\cite{asif2018graspnet} & 33,188 & 15 & 1 & \bbox & \wbox & \bbox & \wbox \\
         & Jacquard~\cite{depierre2018jacquard} & 54,485 & - & 1 & \wbox & \wbox & \bbox & \wbox \\
         & OCID~\cite{ainetter2021end, suchi2019easylabel} & - & - & 1 & \bbox & \wbox & \bbox & \wbox \\
         \midrule
         \multirow{3}{*}{\textit{HOIS}}
         & EPIC-Kitchens~\cite{damen2018scaling} & - & - & 1 & \bbox & \wbox & \wbox & \wbox \\
         & YCB-Affordance~\cite{corona2020ganhand} & 133,936 & 58 & 1 & \bbox & $\raisebox{1pt}{\scalebox{0.5}{\LEFTcircle}}$ & \bbox & \wbox \\ 
         & HO3Pairs~\cite{ye2023affordance} & - & - & 1 & \bbox & \wbox & \wbox & $\raisebox{1pt}{\scalebox{0.5}{\LEFTcircle}}$ \\
         \bottomrule \addlinespace[\belowrulesep]
         \multicolumn{9}{l}{\parbox{\linewidth}{\scriptsize{KEYS -- \# Images:~number of images, OBJ:~number of object categories, AFF:~number of affordance categories, Tran.:~transparency, 3PV:~third person view, HOc:~hand-occlusion, OBJD:~task driven object detection; AFFC:~affordance classification; AFFG:~affordance grounding; HOIS:~hand-object pose estimation and interaction synthesis; GDET:~grasping detection; AFFDS:~affordance detection and segmentation; \bbox: considered, \wbox: not considered, \raisebox{1pt}{\scalebox{0.4}{\LEFTcircle}}: partly considered.}}}
     \end{tabular}
     \label{tab:review_datasets}
     \vspace{-10pt}
 \end{table}

\noindent{\bf Annotations of affordances.}
Previous works proposing datasets either sampled images already available for other tasks such as object detection or image classification~\cite{nguyen2017object,corona2020ganhand,sawatzky2017weakly,luo2022learning,hussain2020fpha,apicella2023affordance,demo2vec2018cvpr}, or collected new images~\cite{guo2023handal,myers2015affordance,jiang2022a4t}. 
Target affordances in most of these datasets are \textit{manually labelled}. For example, affordance segmentation requires to label the pixels of the object regions with an affordance category (fine-grained annotation)~\cite{nguyen2017object,sawatzky2017weakly,myers2015affordance,jiang2022a4t,christensen2022learning,chu2019learning,khalifa2022towards}. However, this procedure is time-consuming and subject to errors, such as missing annotations of objects  ~\cite{myers2015affordance,nguyen2017object}, incomplete annotation (presence of holes) or over the object boundaries, due to clutter or small visible regions. To reduce the annotation effort, a \textit{weakly labelling procedure} requires annotators to only label points of interaction and then to apply a Gaussian filter on the image to expand the point annotation~\cite{luo2022learning,demo2vec2018cvpr}. This procedure was used to annotate two datasets for affordance grounding, OPRA~\cite{demo2vec2018cvpr} and AGD20K~\cite{luo2022learning}. The filtering operation however may cause the affordance map to be non-zero also outside the object boundaries. 
Because of ambiguities in the boundaries of visual affordances and fine-grained annotations, datasets size are often limited to few tens of thousands images. Simulators can generate a large number of synthetic or mixed-reality images with \textit{automatic annotations} while varying the illumination conditions and object models~\cite{apicella2023affordance,depierre2018jacquard,christensen2022learning,chu2019learning}. In this case, the annotation effort consists in the design of the simulated environments, the placement of the object models, and the manual labelling of the mesh with the affordance category~\cite{guo2023handal,apicella2023affordance,christensen2022learning,chu2019learning}. Segmentation masks are obtained by ray-tracing the annotated regions on the object mesh into the simulated camera frame~\cite{apicella2023affordance,christensen2022learning,chu2019learning}. A robotic hand grasping the object can be simulated to save the image of the object, the coordinates of the grasping attempts, and the oriented rectangles~\cite{depierre2018jacquard}. However, images generated with a simulator can differ from images captured with a real camera (sim-to-real gap), hindering the generalisation of trained models to real images. 
An alternative to simulators, is a \textit{(semi-)automatic annotation procedure} using off-the-shelf models~\cite{ye2023affordance,guo2023handal,qu2023rio,mur2023multi}. For example, HANDAL~\cite{guo2023handal} was annotated by using BundleSDF~\cite{wen2023bundlesdf} to estimate the 6D pose of the objects in each frame of a video and to reconstruct their CAD models. Then, the handle of the CAD models were annotated with the affordance \textit{graspable}, and projected in the camera frame to obtain the annotation mask. EPIC-AFF~\cite{mur2023multi} annotation procedure associated the action narrations from EPIC-100~\cite{damen2018scaling} with the hand-object interaction points from VISOR~\cite{darkhalil2022epic}, and projected these points in 3D using a depth estimation model~\cite{ranftl2020towards}; finally, COLMAP~\cite{schonberger2016pixelwise} estimated the camera poses and projected interaction points in the same environment point cloud, causing the affordance regions to cover the object, the background, and the arms of the person when projected in the camera frames. When collecting HO3Pairs~\cite{ye2023affordance} to perform the synthesis of visual affordances, annotators segmented the hand and used an image in-painter~\cite{nichol2021glide} to erase the hand holding the objects reconstructing the occluded part of the object. Although this procedure allows to obtain the image showing the unoccluded object, the reconstruction causes the image quality to degrade with the presence of blurred areas, affecting the performance of trained methods. UMD VL and IIT-AFF VL~\cite{qu2024knowledge} complemented the existing UMD and IIT-AFF annotations with two language description of the task: explicit instructions included object category, action verb, and positional relationship (e.g., “Hand me the [object] on the right to [action]”), and implicit instructions omitting specific object category references (e.g., “Hand me something to [action]”).  CLIP~\cite{radford2021learning} predicted the "[object]" category from the crops of detected objects, while GPT-2~\cite{radford2019language} generated the "[action]" from the template of the instruction to fill. Overall, using off-the-shelf methods can speed-up the labelling procedure, potentially scaling the size of annotated datasets~\cite{deng2024coconut}, requiring annotators to setup the annotation pipeline and check for potential mistakes in generated labels.  

\begin{figure}[t!]
    \centering
    \includegraphics[width=\linewidth]{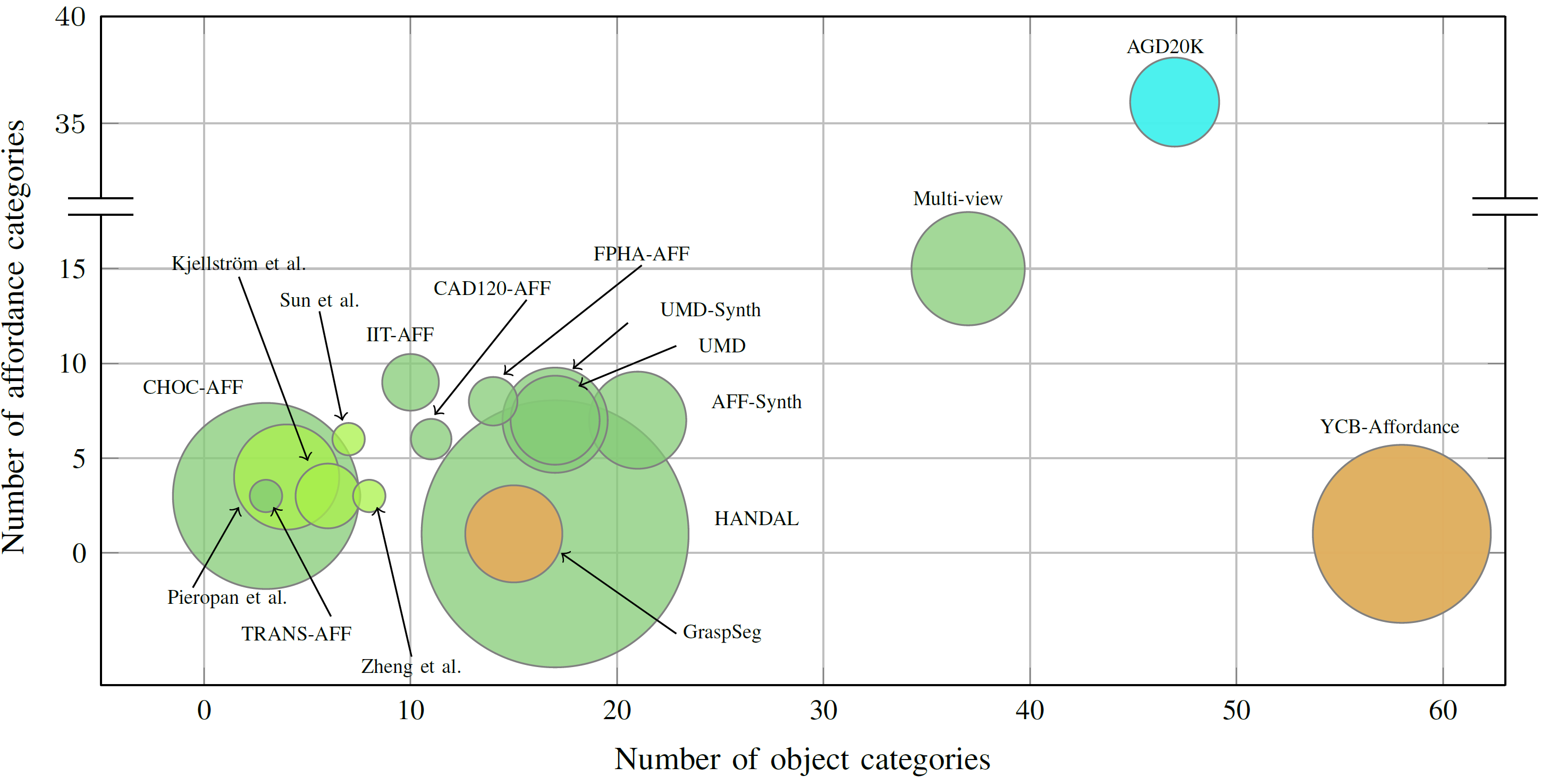}
    \caption{Visualisation of datasets size based on number of images, number of affordance categories, and number of object categories. KEY: 
    \protect\raisebox{2pt}{\protect\tikz \protect\draw[ac,line width=3] (0,0) -- (0.3,0);}~Affordance classification, 
    \protect\raisebox{2pt}{\protect\tikz \protect\draw[ads,line width=3] (0,0) -- (0.3,0);}~Affordance detection and segmentation,
    \protect\raisebox{2pt}{\protect\tikz \protect\draw[ag,line width=3] (0,0) -- (0.3,0);}~Affordance grounding,
    \protect\raisebox{2pt}{\protect\tikz \protect\draw[epe,line width=3] (0,0) -- (0.3,0);}~Hand-object pose estimation.}
    \label{fig:object_affordance_scale}
    \vspace{-10pt}
\end{figure}

\noindent{\bf Camera viewpoint: 3rd and 1st person view.}
The majority of datasets for visual affordance focuses on third person view~\cite{nguyen2017object,guo2023handal,myers2015affordance,luo2022learning,jiang2022a4t,demo2vec2018cvpr,asif2018graspnet,depierre2018jacquard,christensen2022learning,khalifa2022towards,jiang2011efficient}. The camera has a fixed pose and is not mounted on the agent, capturing objects from a constant distance and in a static scene. In some cases~\cite{asif2018graspnet,depierre2018jacquard,jiang2011efficient}, the camera is placed in a top-down view to observe the area close to the agent. These conditions can limit the generalisation of trained models to other scenarios, e.g. different camera view. The first person perspective (egocentric view) includes additional challenges, such as self-occlusions due to the presence of parts of the agent in the collected frames and blur due to the camera movement~\cite{ye2023affordance,sun2010learning,hussain2020fpha,damen2018scaling}. For example, arms are observed from the bottom of an image resulting in objects highly occluded by the hands (e.g. FPHA-AFF~\cite{hussain2020fpha}), or images are affected by blur while people interact with ingredients in a kitchen environment (e.g. EPIC-Kitchens~\cite{ye2023affordance,mur2023multi,damen2018scaling}). Because of these challenges, models trained on egocentric-view datasets might not generalise to third person perspective and vice versa.

\noindent{\bf Occlusions.} 
Most of the datasets~\cite{ye2023affordance,myers2015affordance,asif2018graspnet,depierre2018jacquard,chu2019learning,khalifa2022towards,jiang2011efficient,lakani2019towards} focus on one \textit{unoccluded} object placed on a flat surface (e.g. tabletop or floor) with a fixed setup varying object categories and objects instances. For example, UMD~\cite{myers2015affordance} and Multi-View~\cite{khalifa2022towards} collected more than 15 object categories and annotated more than 5 affordance classes, while controlling the environmental conditions: objects are placed on a rotating table, with the same illumination and background. However, this simple and controlled setup limits the generalisation of models to environments with different illumination and backgrounds, or where multiple objects are present in the scene. 
Only some of the datasets~\cite{nguyen2017object,guo2023handal,corona2020ganhand,sawatzky2017weakly,luo2022learning,apicella2023affordance} contain occlusions caused by clutter in the scenes or human hands holding the objects (hand-occlusions). When objects are occluded, only some of their regions are visible, increasing the difficulty in perceiving the affordances. Hand-occlusion is a main challenge in human-robot collaborations, as erroneous or inaccurate affordance predictions lead to unintended interactions with the object, potentially causing harm to the person (\textit{human safety})~\cite{pang2024stereo,apicella2023affordance}.

\noindent{\bf Objects of interest.} 
In previous works~\cite{nguyen2017object,guo2023handal,corona2020ganhand,sawatzky2017weakly,myers2015affordance,luo2022learning,jiang2022a4t,asif2018graspnet,khalifa2022towards,jiang2011efficient}, the objects most suitable to accomplish a task were considered as objects of interest, and were annotated with the corresponding affordances. 
The majority of these datasets~\cite{pieropan2013functional,nguyen2017object,guo2023handal,sawatzky2017weakly,myers2015affordance,zheng2018high,Kjellstrom2011visual,hussain2020fpha} has fewer than 20 object categories and 10 affordance categories (see Fig.~\ref{fig:object_affordance_scale}), and focuses on the affordances of tools and containers.
Tools are usually opaque and rigid, are used in a kitchen environment (e.g., pan, fork, turner) or for carpentry (e.g., hammer, shovel, saw), and consist of a graspable handle~\cite{nguyen2017object,guo2023handal,myers2015affordance}. 
Compared to tools, perceiving the affordance of containers (e.g. box, cup, glasses) is more challenging, since their properties can change during a manipulation (e.g. the appearance in case of transparent material filled with opaque content)~\cite{apicella2023affordance,sanchez2020benchmark,apicella2022container}. Even if a lot of containers we use in everyday life are transparent, this property is considered only in a few datasets~\cite{nguyen2017object,jiang2022a4t,apicella2023affordance,luo2021one}.

Diversifying the object categories, degrees of occlusions and object poses in datasets is fundamental to tackle the generalisation problem. The generalisation to diverse conditions is relevant in human-robot collaboration and assistive applications, where the environment is not necessarily controlled. 

\section{Affordance prediction reproducibility}
\label{sec:evaluation}

We discuss the evaluation of affordance prediction, focusing on the reproducibility\footnote{Principle of obtaining the same results given the same conditions (i.e. data, training and testing setups, and trained model)~\cite{pineau2021improving}.} issues of current benchmarks, resulting in unfair and inconsistent comparisons. 
Reproducibility allows fair comparisons across methods and helps build upon previous works while understanding their limitations. We then highlight Open Science practices for fair benchmarking. 

\subsection{Reproducibility challenges}

Reproducibility challenges (RCs) in different redefinitions of visual affordance prediction include~\cite{apicella2024segmenting}: 
\begin{enumerate}
    \item data availability for benchmarking (RC1);
    \item availability of a method's implementation (RC2);
    \item availability of trained models (RC3);
    \item details of experimental setups (RC4); and
    \item details of performance measures for evaluation (RC5). 
\end{enumerate}  

In visual affordance prediction, no dataset is collected exclusively for benchmarking methods under specific conditions, such as illumination, clutter, or hand-occlusion (RC1). 
The majority of previous works~\cite{guo2023handal,myers2015affordance,nguyen2016detecting,apicella2023affordance,khalifa2022towards} trained methods on a dataset training split and compared their performance on the testing split of one or more datasets.
Cross-dataset evaluations are mostly avoided due to partial overlapping of affordance classes or of object categories, across the selected datasets~\cite{apicella2024segmenting}. For example, datasets such as UMD~\cite{myers2015affordance}, IIT-AFF~\cite{nguyen2017object}, and Multi-View~\cite{lakani2019towards} share some of the object and affordance classes, but labelled with different conventions, making the comparison of models trained on different dataset difficult. As a consequence, researchers train multiple version of the same model, adapted to the classes of a specific dataset.
Additional documentation, such as metadata, help researchers train or evaluate methods only on common categories by re-ordering them. Moreover, relying only on a single benchmark can lead to limited and not generalisable considerations on model rankings. For example, images in UMD and Multi-View are collected in a laboratory environment with static conditions, such as a fixed camera oriented towards a table where an object is placed (camera-object distance is almost always the same)~\cite{myers2015affordance, khalifa2022towards}. However, in real scenarios the camera might be closer or farther from objects compared to the training setting, hence the performance on the benchmark might not reflect the performance on a real use case. 

The lack of publicly available implementation of methods (RC2)~\cite{zhang2022multi,gu2021visual,yin2022object,zhao2020object}, the lack of publicly available trained models (RC3)~\cite{nguyen2017object,nguyen2016detecting,zhang2022multi,gu2021visual,yin2022new,yin2022object,zhao2020object}, and the lack of details of experimental setups (RC4)~\cite{do2018affordancenet,nguyen2016detecting,zhang2022multi,gu2021visual,yin2022object,zhao2020object}   
can challenge researchers in reproducing previous works for comparative evaluations. 
The release of the model trained weights, of the method and inference pipeline implementation, is a crucial aspect for reproducibility, especially for deep-learning based models, allowing other researchers to test models on their own data without re-training. The availability of model implementation and weights is important when researchers need a comparison, as re-training the model can be too time- and resource-consuming. In case a new dataset is proposed and a previous method needs re-training, only the method implementation is sufficient. The re-implementation of methods and setup is time-consuming and prone to errors, and not always leads to the expected outcome (i.e. results are not replicable or findings are not reproducible). To avoid this issue and to save time, researchers report the results from previous works~\cite{zhang2022multi,gu2021visual,yin2022object,zhao2020object}, resulting in unfair comparisons if the experimental conditions are not the same, and in misleading findings and conclusions.

\begin{table}[t!]
    \centering
    \scriptsize
    \setlength\tabcolsep{2pt}
    \caption{Comparison of training/testing setups used by different methods for affordance detection and segmentation on the UMD dataset~\cite{myers2015affordance}. 
    Due to the setup inconsistencies, direct comparison among models performance is unfair.
    }
    \vspace{-5pt}
    \begin{tabular}{c c cccc cc} 
    \toprule
    \textbf{Training setup} & \textbf{Resolution} & \multicolumn{4}{c}{\textbf{Data augmentation}} & \multicolumn{2}{c}{\textbf{Image resize}} 
    \\
    \cmidrule(lr){3-6}\cmidrule(lr){7-8}
    & & FLIP & SCAL & ROT & JIT & Train. & Test.\\
    \midrule
    AffordanceNet~\cite{do2018affordancenet} & $1000 \times 600$ & $\circ$ & $\circ$ & $\circ$ & $\circ$ & UNK & UNK 
    \\ 
    CNN~\cite{nguyen2016detecting} & $320 \times 240$ & $\circ$ & $\circ$ & $\circ$ & $\circ$ & CC & SLW 
    \\    
    DRNAtt~\cite{gu2021visual} & $320 \times 240$ & $\circ$ & $\circ$ & $\circ$ & $\circ$ & CC & UNK 
    \\ 
    RANet~\cite{zhao2020object} & $224 \times 224$ & $\circ$ & $\circ$ & $\circ$ & $\circ$ & CC & UNK 
    \\
    GSE~\cite{zhang2022multi} & $400 \times 400$ & $\bullet$ & $\bullet$ & $\circ$ & $\circ$ & crop & UNK 
    \\    
    BPN~\cite{yin2022object} & $1000 \times 600$  & $\bullet$ & $\bullet$ & $\bullet$ & $\bullet$ & UNK & UNK 
    \\
    \bottomrule
    \addlinespace[\belowrulesep]
    \multicolumn{8}{l}{\parbox{0.9\linewidth}{\scriptsize{KEYS -- $\bullet$:~considered, $\circ$:~not considered, FLIP:~flipping, SCALE:~scaling, ROT:~rotating, JIT:~colour jittering, Train.:~training set, Test.:~testing set, UNK:~unknown, cc:~centre-crop, SLW:~sliding window}}}\\
    \end{tabular}
    \label{tab:umdtrainingsetup_pre}
    \vspace{-10pt}
\end{table} 

Using the same \textit{experimental setup} to train and test affordance models allows a fair comparison enabling the validation of the technical contributions proposed by a novel work. When releasing the training and testing code is not possible, reporting all details to reproduce a setup becomes fundamental, enabling other researchers to re-implement the setup and correctly compare their solution. 
The experimental setup details include training hyper-parameter values, chosen data splits, image pre-processing (normalisation and cropping procedures), and post-processing. The lack of details of the experimental setup causes methods for affordance detection and segmentation to be often not reproducible~\cite{do2018affordancenet,nguyen2016detecting,zhang2022multi,yin2022object,gu2021visual,zhao2020object}. 
For example, AffordanceNet and BPN do not include image resize during training and testing phases~\cite{do2018affordancenet, yin2022object}, whereas DRNAtt, RANet, and GSE do not include these details during the testing phase. Other details often omitted are the parameters of the optimizers used during training~\cite{gu2021visual,qian2024affordancellm,li2024one,qian2023understanding}. 
Apicella et al.'s work~\cite{apicella2024segmenting} showed that the lack of details in the experimental setup led to unfair and inconsistent comparisons. 

Previous works evaluated the performance of different methods using scores or metrics to quantify the discrepancy between predictions and annotations (more details in Supp. Mat.). Describing a performance measure help other researchers understand if the experiment validates their claim or if a different measure should be chosen. Providing the mathematical formulation of the scores helps disambiguate similar meaning but different implementations, especially when a public evaluation toolkit is not used or referred to. For example,  mean \textit{IoU} can be the average of all the \textit{IoU}s between prediction and annotation, or the \textit{IoU} considering the full set of predictions and annotations. 
Previous works evaluated a few methods with different performance measures or datasets, making comparison and ranking not possible. For example, the performance of  AdaptiveNet~\cite{sawatzky2017adaptive} and STRAP~\cite{cui2023strap} was compared on CAD120-AFF using IoU, instead of UMD using $F^w_\beta$ as most of available methods.

Affordance detection and segmentation methods are difficult to reproduce due to missing implementation and lack of setups details~\cite{do2018affordancenet,nguyen2016detecting,zhang2022multi,gu2021visual,yin2022object,zhao2020object}. We report the training and testing setups of affordance detection and segmentation methods on the UMD dataset in 
Table~\ref{tab:umdtrainingsetup_pre}. 
Despite being trained and tested on the same dataset, models' performance is not directly comparable due to inconsistencies in the setups such as the image resize procedure (image cropping or input resolution) and augmentation procedure during training.  

Inconsistencies can also be present in previous methods adapted into a baseline to compare with. For example, due to the missing annotation of the object pose in the training/testing dataset, AffordanceDiffusion~\cite{ye2023affordance} is compared with the coarse hand prediction of GanHand~\cite{corona2020ganhand}. 
However, since a part of the architecture and of the training procedure is missing, the result is only a proxy to the (unknown) performance of GanHand.

\begin{table}[t!]
    \centering
    \setlength\tabcolsep{2pt}
    \scriptsize
    \caption{Affordance Sheet, inspired by Model Cards~\cite{mitchell2019model}, to favour transparency and reproducibility of works for visual affordance predictions conditioned on robotic tasks. Example filled with ACANet~\cite{apicella2023affordance} details.}
    \vspace{-5pt}
    \begin{tabular}{|l |c c c c c c|}
    \hline
    \multicolumn{7}{|c|}{\textbf{ACANet}} \\
    \hline
    \textbf{Affordance task}
     & OBJL & FUNC & FUNS & EPE & EIS & \\
     & \wbox & \wbox & \bbox & \wbox & \wbox & \\
    \hline
    \parbox{0.28\columnwidth}{\textbf{Datasets\\(RC1)}}
    & \multicolumn{1}{l}{\textit{Name:}} & \multicolumn{5}{l|}{CHOC-AFF} \\ 
    & \multicolumn{1}{l}{\textit{Record link*:}} & \multicolumn{5}{l|}{\url{https://doi.org/10.5281/zenodo.5085800}} \\
    & \multicolumn{1}{l}{\textit{Licence:}} & \multicolumn{5}{l|}{\parbox{0.4\columnwidth}{\strut CC BY 4.0 \strut}}\\  
    \hline
    \parbox{0.28\columnwidth}{\textbf{Proposed method \\ (RC2, RC3)}} & \multicolumn{1}{l}{\textit{Record link*:}} & \multicolumn{5}{l|}{\url{https://doi.org/10.5281/zenodo.8364196}} \\ 
    & \multicolumn{1}{l}{\textit{Code link:}} & \multicolumn{5}{l|}{\url{https://github.com/apicis/aff-seg/}} \\
    & \multicolumn{1}{l}{ \textit{Model card}:} & \multicolumn{5}{l|}{\bbox} \\
    & \multicolumn{1}{l}{\textit{Licence:}} & \multicolumn{5}{l|}{CC BY-NC-SA 4.0} \\  
    \hline
    \parbox{0.28\columnwidth}{\textbf{Experimental setup\\ (RC4)}} & \multicolumn{6}{l|}{\textit{Data splits:}} \\
    & &  \multicolumn{5}{l|}{ 
    \begin{tabular}{l c}
    \toprule
    Set & Images \\
    \midrule
    Training & 89,856 \\
    Validation & 17,280 \\
    Testing 1 & 13,824 \\
    Testing 2 & 17,280 \\
    \bottomrule
    \end{tabular}}
    \\
    & \multicolumn{6}{l|}{\textit{Hyperparameters:}} \\
    & & \multicolumn{5}{l|}{
    \begin{tabular}{l c}
    \toprule
    Name & Value \\
    \midrule
    batch size & 2 \\
    learning rate & 0.001 \\
    schedule & 0.5x \\
    patience & 3 \\ 
    optmizer & SGD \\
    momentum & 0.9  \\ 
    weight decay & 0.0001 \\ 
    resize & [1, 1.5] \\
    flip & 0.5 \\
    \bottomrule
    \end{tabular}
    }\\
    & \multicolumn{1}{l}{\textit{Resize procedure:}} & \multicolumn{5}{l|}{center crop  $480 \times 480$}\\  
    \hline
    \parbox{0.28\columnwidth}{\textbf{Performance measures\\(RC5)}} & \multicolumn{1}{l}{\textit{Description:}} & \multicolumn{5}{l|}{\parbox{0.4\columnwidth}{\strut Per-class Jaccard index measures the overlap between predicted and annotated segmentation masks, and quantifies how much they are similar in size \strut}}\\  
    & \multicolumn{1}{l}{\textit{Formulation:}} & \multicolumn{5}{l|}{\parbox{0.4\linewidth}{
    $\frac{\sum_{n=1}^{N} \sum_{\boldsymbol{y} \in I_n} TP^{\boldsymbol{y}}_n}{ \sum_{n=1}^{N} \sum_{\boldsymbol{y} \in I_n} TP^{\boldsymbol{y}}_n + FP^{\boldsymbol{y}}_n + FN^{\boldsymbol{y}}_n} \: $}}\\  
    & \multicolumn{1}{l}{\textit{Limitations:}} & \multicolumn{5}{l|}{\parbox{0.4\linewidth}{\strut Jaccard Index does not consider the masks shape \strut}} \\    
    \hline
    \textbf{Robot validation} 
    & \multicolumn{1}{l}{\textit{Robot model:}} & \multicolumn{5}{l|}{\parbox{0.35\columnwidth}{-}}\\
    & \multicolumn{1}{l}{\textit{End-effector:}} & \multicolumn{5}{l|}{\parbox{0.35\columnwidth}{-}}\\
    & \multicolumn{1}{l}{\textit{Experiment:}} & \multicolumn{5}{l|}{\parbox{0.35\columnwidth}{-}}\\
    \hline
    \multicolumn{7}{|l|}{\parbox{0.98\columnwidth}{\scriptsize{\strut \textit{Legend}: OBJL:~object localisation; FUNC:~functional classification; FUNS:~functional segmentation; EPE:~hand pose estimation; EIS:~hand interaction synthesis;
    RC:~reproducibility challenge.\\
    \textbf{Notes}: *data and weights of the trained model are recommended to be placed in a repository that favours long-term persistence and accessibility.
    \strut}}} \\
    \hline
    \end{tabular}
    \label{tab:affsheet}
    \vspace{-8pt}
\end{table}


The redefinition of the visual affordance problem (see Sec.~\ref{sec:related_works}) can also result in experimental validations ignoring datasets and benchmarks of partially overlapping formulations. For example, works on affordance grounding~\cite{luo2022learning,qian2024affordancellm,li2024one,cuttano2024does} do not compare the performance of proposed methods with that of affordance segmentation methods~\cite{nguyen2016detecting,zhang2022multi,gu2021visual,zhao2020object}, even if the problem formulation is similar~\cite{luo2022learning, nguyen2016detecting}. Methods for affordance segmentation output a binary mask for each action in a predefined set of classes, whereas methods for affordance grounding output a confidence map describing where an action known a priori can take place in the image. Despite these differences, comparing methods for both affordance grounding and affordance detection and segmentation can explain if using action as input (affordance grounding) to a model provides any advantage. 

\subsection{In support of reproducibility: Affordance Sheets}
To promote reproducibility in affordance prediction, we propose the Affordance Sheet, an organised collection of good practices favouring fair comparisons and the development of new solutions (see Table~\ref{tab:affsheet}). Model cards~\cite{mitchell2019model} were previously introduced to improve the methods transparency and to raise awareness about limitations, by describing the method, the experimental setup, and the applications or conditions leading to underperformance. Our Affordance Sheet integrates Model Cards complementing the released information. 

The first section identifies which problems the affordance model tackles, helping researchers understand what are the competing methods and assess their performance of solutions under the same inputs and conditions.  
When proposing a new problem partially overlapping with another one, previous models can be used or adapted to validate the method. For example, selecting the channel of an affordance segmentation output based on the action considered by the affordance grounding method enables the comparison between methods for affordance segmentation and methods for affordance grounding. To compare the grounding and segmentation outputs, the grounding confidence map can be converted to a binary mask via thresholding; alternatively, the segmentation map can be converted to a confidence map 
by using Gaussian blur. 

The second section of the Affordance Sheet describes the datasets (RC1) used by the proposed solution, to detail their characteristics, share the link to the data, and the license informing about data permissions. We recommend future benchmarks to also release a detailed description on how to use and visualize data so that researchers can get acquainted with the format. Moreover, we recommend that future benchmarks evaluate models under different conditions, such as generalization to different object instances, object categories, object poses, backgrounds, and clutter. Benchmarks of models for tasks different from visual affordance prediction, such as COCO for object detection and instance segmentation~\cite{lin2014microsoft} release only the training and validation sets while keeping a private testing set to not bias the designer of the architecture~\cite{lin2014microsoft}. The availability of a testing set can lead researchers to make changes aimed at improving performance scores rather than formulating contributions that advance the field.

The third section highlights the model characteristics (RC2, RC3) integrating information in the model card (if available). Providing model cards~\cite{mitchell2019model}, along with its implementation and trained weights, helps detail the description of models supporting other researchers to build upon. When not available, we encourage the re-implementation and retraining of the models as a contribution for the community (e.g. a previous work re-implemented, retrained, and released models for affordance detection and segmentation due to the lack of available models~\cite{apicella2024segmenting, apicella2023affordance}). As recommended for datasets, we encourage providing a link to the trained model's weights and a license detailing the allowed uses. Without a license, the code is automatically protected by copyright, hence other researchers can not directly use the method implementation to reproduce results or as baseline, as for some previous works~\cite{hussain2020fpha,cui2023strap,nagarajan2019grounded,qian2023understanding,hadjivelichkov2023one}. 

By providing the details of the experimental setup to train and evaluate methods (RC4), the fourth section of the Affordance Sheet is fundamental to correctly use previous methods and develop a solution under the same conditions.
Setup conditions include pre-processing and post-processing information such as data splits, resize procedures, data normalisation, and hyper-parameters choice. The lack of these details can result in models with significantly different parameters, and hence leading to unfair comparisons with previous works.

The fifth section of the Affordance Sheet focuses on the performance measures (RC5), the criteria used to validate and compare methods with previous solutions. Providing a stand-alone toolkit implementing the performance measures ensures the replicability of the results across different works while including new methods. For visual affordance prediction, we recommend evaluating the performance of models using more than one measure to provide a more comprehensive analysis while identifying different aspects and limitations of the models. For example, in affordance segmentation, precision focuses on how many of the predicted pixels have the correct class and recall emphasizes how many of the annotated pixels are correctly predicted. Therefore, computing more than one score (and avoiding using a single score aggregating multiple performance measures) reduces the risk of drawing misleading conclusions that are based only on partial results.

The last section describes the validation of the method through a robotic setup. In previous works, few of the methods were validated using a robotic platform~\cite{nguyen2017object,do2018affordancenet,yin2022object,tong2024oval,christensen2022learning}. Unlike previous sections of the Affordance Sheet, the robot validation depends on the availability of a robot. When a robot experiment can be performed, we recommend reporting the characteristics of the setup, the robotic hand specifics, and the description of the experiment in terms of object and conditions. This transparent reporting allows researchers to assess methods using a common platform.

\section{Future directions}
\label{sec:future_directions}

In this section we discuss unexplored directions: estimating object physical properties, integration with AI agents, scaling datasets size, and benchmarking models performance.

\noindent{\bf Object physical properties}. 
Relating affordance prediction and estimation of object physical properties is far from easy. Humans have different ways of grasping objects depending on the action they want to perform and how the object properties (e.g. mass) influences the action through the physics of the interaction~\cite{lastrico2023expressing}. Estimating object physical properties only from images might be too complex, and other modalities, such as language, audio, and haptic, could be included in our proposed formulation~\cite{xompero2022corsmal}. Multimodal models have shown better generalisation to novel and different object categories in tasks such as open-vocabulary object detection~\cite{Liu2024ECCV_GroundingDINO} and segmentation~\cite{Wu2024PAMI_OpenVocabulary}. Language can be processed to select the most appropriate grasp for the task~\cite{tulbure2025llm}. Audio could complement the visual modality when the appearance of the object is not reliable, e.g. an opaque container whose content is not visible~\cite{xompero2022audio}. Haptic could provide a feedback on the force that the agent applies on the object~\cite{feng2020center}. 

\noindent{\bf AI agents, human-in-the-loop, and VLA models.}
An AI agent~\cite{durante2024agent} integrating visual affordance requires steps such as understanding (perception), reasoning (relating affordances, objects, and physical properties, conditioned to the task to accomplish), planning (actuation to accomplish the task), and recovering from errors. Learning to predict visual affordances for hand-object interactions can benefit from human demonstrations of the actions to perform, in the same way humans prompt models with examples showing how to solve tasks~\cite{garg2022can}. Our formulation can be extended to include the feedback from a person at different stages (human-in-the-loop)~\cite{yu2024rlhf} to correct the prediction mistakes or also to inject task specific knowledge in the process. Another research direction is the conditioning of end-to-end models with affordance~\cite{bjorck2025gr00t, team2025gemini}. These end-to-end methods do not explicitly model object affordances, and require thousands of demonstration data during training to generalise to different objects. Integrating affordance information coming from our formulation in end-to-end methods can improve spatial reasoning and generalisation to unseen tasks~\cite{yuan2024robopoint}.

\noindent{\bf Scaling visual affordance datasets.}
Datasets cannot be easily re-used across different tasks or for the unified case, as each dataset is specific to an affordance redefinition rather than the unified formulation. Moreover, the annotation of object affordances in images and videos is not trivial due to the unclear boundaries of the region on the object, the overlapping of different actions on the same region, and the difficulty of labelling the agent's hand pose on objects in the scene. These challenges limit the cross-datasets evaluation of methods and the scalability of datasets for visual affordance, as manual annotations are time-consuming and ambiguous, and require expensive resources (as discussed in Sec.~\ref{sec:review_datasets} and Sec.~\ref{sec:evaluation}). To scale the number of training data, datasets having similar annotation could be merged, adjusting the annotation, or adapting previous methods to provide weakly or self-supervised annotation (e.g. HANDAL~\cite{guo2023handal}). The combination of different methods could help using in-the-wild images with objects in challenging poses and with different backgrounds.

\noindent{\bf Benchmarking visual affordance.}
Reproducibility and advancements in the design of novel solutions has been facilitated by available datasets, benchmarks and competitions in various computer vision tasks (e.g. BOP for object 6D pose estimation~\cite{hodan2018bop}). 
However, benchmarks for visual affordance predictions are not yet available. 
Nevertheless, solutions based on our generic formulation and novel methods can be designed for robotic grasping and manipulation tasks~\cite{Sun2024RAM}, picking in clutter~\cite{DAvella2024RAM}, and human-to-robot object handovers~\cite{sanchez2020benchmark}, whose benchmarking protocols and competitions are available. A benchmarking protocol specific to visual affordance could be  designed and included in existing competitions to further promote reproducibility and engagement. 

\bibliographystyle{IEEEtran}
\bibliography{ax_short_strings,refs}

\appendices
\section{Performance measures}
\label{sec:review_measures}

We detail the measures to evaluate the performance of models for visual affordance prediction using our formulation as a reference. For each component or sub-task, more than one performance measure can provide a complete assessment and avoid drawing partial or misleading conclusions (see Table~\ref{tab:performance_measures}). We highlight the characteristics and limitations of performance measures used by previous works.

\noindent{\bf Functional classification.}
The performance measures for assessing functional classification are computed  across all $N$ samples (image, task, hand) of a given dataset, each associated with an affordance category $a$. For each category, a true positive ($TP$) is a sample for which the model predicts $a$ and the annotation is also $a$; a false positive ($FP$) is a sample for which the model predicts $a$, but the annotation is a different class; a false negative ($FN$) is a sample with annotation $a$, but the model predicts a different class; a true negative ($TN$) is a sample for which both the model prediction and the annotation are a class different from $a$.
\textit{Per-class accuracy} ($A$) measures the amount of affordance predictions matching the annotations:
\begin{equation}
    A = \frac{\sum^N_{n=1} TP_n + TN_n}{\sum^N_{n=1} TP_n + TN_n + FP_n + FN_n}.
    \label{eq:acc}
\end{equation}
\textit{Per-class precision} ($P$) measures the amount of class predictions matching the annotations among all class predictions:
\begin{equation}
    P = \frac{\sum^N_{n=1} TP_n}{\sum^N_{n=1} TP_n + FP_n}.
    \label{eq:prec1}
\end{equation}
\textit{Per-class recall} ($R$) measures the amount of class predictions matching the annotations among all class annotations:
\begin{equation}
    R = \frac{\sum^N_{n=1} TP_n}{\sum^N_{n=1} TP_n + FN_n}.
    \label{eq:rec1}
\end{equation}
\textit{Per-class F1 score} ($F$) is the harmonic mean of per-class precision and recall:
\begin{equation}
    F = 2\frac{P R}{P + R}.
    \label{eq:f1}
\end{equation}

\begin{table}[t!]
    \scriptsize
    \setlength\tabcolsep{1pt}
    \centering
    \caption{Performance measures to evaluate methods for visual affordance prediction. Highlighted in {\colorbox{lightgray}{grey}} the measures we recommend for evaluation. }
    \begin{tabular}{l c l ccccc}
         \toprule
         \textbf{Performance measure} & \textbf{Variable} & \textbf{Reference} & \textbf{FUNC} & \textbf{FUNS} & \textbf{EPE} & \textbf{EIS} & \textbf{ROBV}\\
         \midrule
         \rowcolor{lightgray} Accuracy & $A$ & Eq.~\ref{eq:acc} & $\bullet$ & $\circ$ & $\circ$ & $\circ$ & $\circ$ \\
         \rowcolor{lightgray} F1 score & $F$ & Eq.~\ref{eq:f1} & $\bullet$ & $\circ$ & $\circ$ & $\circ$ & $\circ$ \\
         \rowcolor{lightgray} Precision & $P$ & Eq.~\ref{eq:prec1}, Eq.~\ref{eq:prec2} & $\bullet$ & $\bullet$ & $\circ$ & $\circ$ & $\circ$ \\
         \rowcolor{lightgray} Recall & $R$ & Eq.~\ref{eq:rec1}, Eq.~\ref{eq:rec2} & $\bullet$ & $\bullet$ & $\circ$ & $\circ$ & $\circ$ \\
         \rowcolor{lightgray} Jaccard index & $J$ & Eq.~\ref{eq:jaccard_aff} & $\circ$ & $\bullet$ & $\circ$ & $\circ$ & $\circ$\\
         Weighted F-score & $F^w_\beta$ & \cite{margolin2014evaluate} & $\circ$ & $\bullet$ & $\circ$ & $\circ$ & $\circ$\\
         Kullback-Leibler Divergence & - & \cite{bylinskii2018different} & $\circ$ & $\bullet$ & $\circ$ & $\circ$ & $\circ$\\
         Similarity & - & \cite{swain1991color} & $\circ$ & $\bullet$ & $\circ$ & $\circ$ & $\circ$\\
         Normalized Scanpath Saliency & - & \cite{peters2005components} & $\circ$ & $\bullet$ & $\circ$ & $\circ$ & $\circ$\\
         \rowcolor{lightgray} Analytical grasp score & - & \cite{ferrari1992planning} & $\circ$ & $\circ$ & $\bullet$ & $\circ$ & $\circ$\\
         \rowcolor{lightgray} Interpenetration volume & - & \cite{corona2020ganhand} & $\circ$ & $\circ$ & $\bullet$ & $\circ$ & $\circ$\\
         Contact fingers & - & \cite{corona2020ganhand} & $\circ$ & $\circ$ & $\bullet$ & $\circ$ & $\circ$\\
         \rowcolor{lightgray} Fréchet Inception Distance & $FID$ & \cite{heusel2017gans}, Eq.~\ref{eq:fid} & $\circ$ & $\circ$ & $\circ$ & $\bullet$ & $\circ$\\
         Contact Recall & - & \cite{ye2023affordance} & $\circ$ & $\circ$ & $\circ$ & $\bullet$ & $\circ$\\
         \rowcolor{lightgray} Success rate & - & - & $\circ$ & $\circ$ & $\circ$ & $\circ$ & $\bullet$\\
         \bottomrule\addlinespace[\belowrulesep]
         \multicolumn{8}{l}{\parbox{\linewidth}{\scriptsize{KEYS -- FUNC:~functional classification; FUNS:~functional segmentation; EPE:~hand pose estimation; EIS:~hand interaction synthesis; ROBV:~robot validation; $\bullet$: considered, $\circ$: not considered.}}}
     \end{tabular}
     \label{tab:performance_measures}
     \vspace{-5pt}
 \end{table}

When evaluating the performance of affordance classification methods, previous works~\cite{pieropan2013functional,sun2010learning,zheng2018high,Kjellstrom2011visual} showed confusion matrices and accuracy. However, the level of detail of confusion matrices makes it difficult to quantitatively compare methods. For datasets with imbalanced classes, accuracy is misleading because a high value can be obtained by predicting always the most frequent class. On the contrary, using precision, recall, and F1 provides a complementary analysis while considering imbalanced classes, because precision focuses on false positives and recall on false negatives. 

\noindent{\bf Functional segmentation.} 
The performance measures for assessing the functional segmentation of are \textit{per-class precision} ($P$), \textit{per-class recall} ($R$) and \textit{per-class Jaccard index} ($J$) or \textit{Intersection over Union} ($IoU$). To compute these measures, the output probability maps of the model $[0, 1]^{W \times H}$ are converted into integer values $\{0, 1\}^{W \times H}$ for example using a threshold. As for functional classification, true positives ($TP$), false positives ($FP$), and false negatives ($FN$) are defined for each class $a$. Given the model prediction $\hat{S}$ and the segmentation annotation of the image $S$, a true positive is a pixel $\boldsymbol{y} \in I_n$ that is predicted as 1 in $\hat{S}_{n}$ and the corresponding pixel in $S_{n}$ is annotated as 1; a false positive is a pixel $\boldsymbol{y} \in I_n$ that is predicted as 1 in $\hat{S}_{n}$ but annotated as 0 in $S_{n}$; a false negative is a pixel $\boldsymbol{y} \in I_n$ that is predicted as 0 in $\hat{S}_{n}$, but the corresponding pixel in $S_n$ is annotated as 1. 
\textit{Per-class precision} measures the percentage of true positives among all positive predicted pixels, 
\begin{equation}
   P = \frac{\sum_{n=1}^{N} \sum_{\boldsymbol{y} \in I_n} TP^{\boldsymbol{y}}_n}{ \sum_{n=1}^{N} \sum_{\boldsymbol{y} \in I_n}  TP^{\boldsymbol{y}}_n + FP^{\boldsymbol{y}}_n}.
   \label{eq:prec2}
\end{equation} 
\textit{Per-class recall} measures the percentage of true positive pixels with respect to the total number of positive pixels,
\begin{equation}
   R = \frac{\sum_{n=1}^{N} \sum_{\boldsymbol{y} \in I_n} TP^{\boldsymbol{y}}_n}{ \sum_{n=1}^{N} \sum_{\boldsymbol{y} \in I_n} TP^{\boldsymbol{y}}_n + FN^{\boldsymbol{y}}_n} \:.
   \label{eq:rec2}
\end{equation}
\textit{Per-class Jaccard index} combines precision and recall measuring the overlap between predicted and annotated segmentation masks, and quantifying how much they are similar in size,
\begin{equation}
   J = \frac{\sum_{n=1}^{N} \sum_{\boldsymbol{y} \in I_n} TP^{\boldsymbol{y}}_n}{ \sum_{n=1}^{N} \sum_{\boldsymbol{y} \in I_n} TP^{\boldsymbol{y}}_n + FP^{\boldsymbol{y}}_n + FN^{\boldsymbol{y}}_n} \:.
   \label{eq:jaccard_aff}
\end{equation}
We recommend to report the Jaccard index with complementary performance scores, such as precision and recall, to provide a more comprehensive evaluation and insights.

Most affordance detection and segmentation works~\cite{do2018affordancenet,nguyen2017object,sawatzky2017weakly,hussain2020fpha,nguyen2016detecting,yin2022object,gu2021visual,yin2022new,zhao2020object,zhang2022multi,chen2023adosmnet} evaluated the performance of methods using the \textit{weighted F-score} ($F^{w}_{\beta}$)~\cite{margolin2014evaluate}. $F^{w}_{\beta}$ weighs false positives based on the Euclidean distance to the closest annotated pixels, ignoring the classes that are not in the annotated mask. To compare the predicted probability map with the annotation, affordance grounding works~\cite{luo2022learning,qian2024affordancellm,li2024one,cuttano2024does} used \textit{Kullback–Leibler divergence}~\cite{bylinskii2018different}, \textit{Similarity}~\cite{swain1991color}, \textit{Normalized Scanpath Saliency}~\cite{peters2005components}. 
The \textit{Kullback-Leibler Divergence} gives more importance to false negatives compared to false positives. In particular, a false positive results in a \textit{Kullback-Leibler Divergence} value close to 0, whereas a false negative can cause the value to be high (potentially infinite). \textit{Similarity} combines together the information of false positives and false negatives, assigning a low value to both errors and hence resulting in an ambiguous interpretation. \textit{Normalized Scanpath Saliency} considers the prediction values around a neighbourhood of the annotated points. This measure can lead to misleading insights, since the false positives outside the annotation neighbourhood are discarded.

\noindent{\bf Hand pose estimation and synthesis.} Predicted poses of the hand, also different from the annotated ones, can enable a robot to complete the task, making the evaluation of estimated and synthesised hand poses challenging. We therefore recommend using \textit{interpenetration} and \textit{analytical grasp score} to evaluate the estimated pose, and  \textit{Fréchet Inception Distance} to evaluate the synthesised pose.

The \textit{interpenetration}~\cite{corona2020ganhand} is the volume in common between object and hand voxels representation (the lower the better). The measure does not consider if the predicted pose is not feasible and cannot be computed if the datasets lacks the annotation of the object pose or the annotation of thehand pose.
\textit{Analytical grasp score}~\cite{ferrari1992planning} computes an approximation of the minimum force to be applied to break the grasp stability by solving a quadratic program. The minimum force corresponds to the smallest Euclidean distance from the origin to any point inside the convex hull composed by all feasible forces and torques combinations.
To evaluate the hand pose, Corona et al.~\cite{corona2020ganhand} also used the \textit{average number of contact fingers}: the higher the number of fingers in contact, the stronger the grasp. This measure, however, can penalise actions or objects for which the number of contact fingers is low (e.g. when grasping a glass from the stem).
\textit{Fréchet Inception Distance} ($FID$)~\cite{heusel2017gans} quantifies the similarity between two Gaussian distributions, one fitted on the synthesised images $\hat{G} \sim (\hat{\mu}, \hat{C})$ (where $\mu$ is the mean and $\hat{C}$ the covariance) and the other on the testing set images $G \sim (\mu, C)$ (or ground truth). In particular, the two Gaussian distributions are fitted on the Inception feature representations~\cite{szegedy2015going}. $FID$ is computed as:
\begin{equation}
    FID = ||\mu - \hat{\mu}||^2_2 + Tr(C + \hat{C} - 2 (C\hat{C})^{\frac{1}{2}}) \:,
    \label{eq:fid}
\end{equation}
where $Tr$ is the trace operator (i.e. the sum of the diagonal elements of a matrix). The first term, $||\mu - \hat{\mu}||^2_2$, measures the squared difference between the means of the real and generated distributions. A smaller difference indicates that the generated and real images have similar overall features. The second term, $Tr(C + \hat{C} - 2 (C\hat{C})^{\frac{1}{2}})$, compares the covariances of the real and generated distributions (diversity). A low $FID$ score implies high similarity between the generated images distribution and the testing ones. A high $FID$ score suggests that the distribution of the generated images differs from the distribution of the testing images, either in terms of overall features (mean) or diversity of features (covariance). To evaluate AffordanceDiffusion, Ye et al.~\cite{ye2023affordance} also compute \textit{contact recall} that is the amount of generated hands classified as ``in-contact'' with the object in the image by an off-the-shelf method~\cite{shan2020understanding}. However, in case of unseen objects or unseen conditions (illumination, colour of the background), the method could misclassify whether the hands are in contact or not, leading to a mistake in the computation of \textit{contact recall}. 

\noindent{\bf Overall evaluation.} 
If a robot is available, models performance can be assessed in real conditions using success rate~\cite{do2018affordancenet,yin2022new,yin2022object}. Reproducing experiments based on success rate is difficult for some tasks and requires a rigorous protocol. The setup should include information on the object instances, robot model, software versions, and relative poses between object and robot. The evaluation should consider separately if actions are successful (e.g., grasping and lifting), also waiting a fixed amount of time to check if the object falls. When the task is part of other benchmarks~\cite{sanchez2020benchmark}, using the available performance measures enriches the evaluation. 

\newpage
\section{Affordance sheets}
We provide some examples of compiled affordance sheets for related works based on the available information~\cite{chen2024vltp,nagarajan2020ego,lundell2021multi}. In particular, VLTP~\cite{chen2024vltp} is a method for object localisation, EgoTopo~\cite{nagarajan2020ego} for functional classification, and Multi-FinGAN~\cite{lundell2021multi} for Hand pose estimation.

\begin{table}[H]
    \centering
    \setlength\tabcolsep{2pt}
    \scriptsize
    \caption{Affordance Sheet, inspired by Model Cards~\cite{mitchell2019model}, to favour transparency and reproducibility of works for visual affordance predictions conditioned on robotic tasks. Example filled with Multi-FinGAN~\cite{lundell2021multi} details. }
    \begin{tabular}{|l |c c c c c c|}
    \hline
    \multicolumn{7}{|c|}{\textbf{Multi-FinGAN}} \\
    \hline
    \textbf{Affordance task}
     & OBJL & FUNC & FUNS & EPE & EIS & \\
     & \wbox & \wbox & \wbox & \bbox & \wbox & \\
    \hline
    \parbox{0.28\columnwidth}{\textbf{Datasets\\(RC1)}}
    & \multicolumn{1}{l}{\textit{Name:}} & \multicolumn{5}{l|}{-} \\ 
    & \multicolumn{1}{l}{\textit{Record link*:}} & \multicolumn{5}{l|}{\parbox{0.4\columnwidth}{\url{https://github.com/aalto-intelligent-robotics/Multi-FinGAN/blob/main/data/download_train_data.sh}}} \\
    & \multicolumn{1}{l}{\textit{Licence:}} & \multicolumn{5}{l|}{\parbox{0.4\columnwidth}{\strut - \strut}}\\  
    \hline
    \parbox{0.28\columnwidth}{\textbf{Proposed method \\ (RC2, RC3)}} & \multicolumn{1}{l}{\textit{Record link*:}} & \multicolumn{5}{l|}{\parbox{0.4\columnwidth}{\url{https://drive.google.com/file/d/19462M8s3tEXe_1_riHuvQegLxzdX-kl2/view}}} \\ 
    & \multicolumn{1}{l}{\textit{Code link:}} & \multicolumn{5}{l|}{\parbox{0.4\columnwidth}{\url{https://github.com/aalto-intelligent-robotics/Multi-FinGAN}}} \\
    & \multicolumn{1}{l}{ \textit{Model card}:} & \multicolumn{5}{l|}{\wbox} \\
    & \multicolumn{1}{l}{\textit{Licence:}} & \multicolumn{5}{l|}{MIT} \\  
    \hline
    \parbox{0.28\columnwidth}{\textbf{Experimental setup\\ (RC4)}} & \multicolumn{6}{l|}{\textit{Data splits:}} \\
    & & \multicolumn{5}{l|}{ 
    \begin{tabular}{l c}
    \toprule
    Set & Images \\
    \midrule
    Training & 3000 \\
    Validation & - \\
    Testing & - \\
    \bottomrule
    \end{tabular}}
    \\
    & \multicolumn{6}{l|}{\textit{Hyperparameters:}} \\
    & & \multicolumn{5}{l|}{ 
    \begin{tabular}{l c}
    \toprule
    Name & Value \\
    \midrule
    batch size & 100 \\
    learning rate & 0.0001 \\
    schedule & linear after 400 epochs \\
    patience & - \\ 
    optmizer & Adam \\
    momentum & default  \\ 
    weight decay & default \\ 
    resize & - \\
    flip & - \\
    \bottomrule
    \end{tabular}}
    \\
    & \multicolumn{1}{l}{\textit{Resize procedure:}} & \multicolumn{5}{l|}{Object centric crops resized to $256 \times 256$}\\  
    \hline
    \parbox{0.28\columnwidth}{\textbf{Performance measures\\(RC5)}} & \multicolumn{1}{l}{\textit{Description:}} & \multicolumn{5}{l|}{\parbox{0.4\columnwidth}{Interpenetration: amount of voxels in common between object and end-effector. \strut}}\\  
    & \multicolumn{1}{l}{\textit{Formulation:}} & \multicolumn{5}{l|}{\parbox{0.4\linewidth}{
    -}}\\  
    & \multicolumn{1}{l}{\textit{Limitations:}} & \multicolumn{5}{l|}{\parbox{0.4\linewidth}{\strut Interpenetration does not take into account the predicted pose feasibility. \strut}} \\    
    \hline
    \textbf{Robot validation} 
    & \multicolumn{1}{l}{\textit{Robot model:}} & \multicolumn{5}{l|}{\parbox{0.35\columnwidth}{Franka Emika Panda}}\\
    & \multicolumn{1}{l}{\textit{End-effector:}} & \multicolumn{5}{l|}{\parbox{0.35\columnwidth}{Barrett hand}}\\
    & \multicolumn{1}{l}{\textit{Experiment:}} & \multicolumn{5}{l|}{\parbox{0.4\columnwidth}{Intel RealSense D435 camera looking at the scene at 45 degree viewpoint. The model generates 20 grasps per object and then intersection and quality metric of each grasp are computed. The first physically reachable grasp with lowest intersection and highest quality metric is executed on the real robot. The robot needs to grasp the object and, without dropping it, move to the start position and rotate the hand ±90° around the last joint (success). If the object was dropped during the manipulation, the grasp is considered unsuccessful.}}\\
    \hline
    \multicolumn{7}{|l|}{\parbox{0.98\columnwidth}{\scriptsize{\strut \textit{Legend}: OBJL:~object localisation; FUNC:~functional classification; FUNS:~functional segmentation; EPE:~hand pose estimation; EIS:~hand interaction synthesis;
    RC:~reproducibility challenge; '-': information not available.\\
    \textbf{Notes}: *data and weights of the trained model are recommended to be placed in a repository that favours long-term persistence and accessibility.
    \strut}}} \\
    \hline
    \end{tabular}
    \label{tab:affsheet}
    \vspace{-5pt}
\end{table}
\begin{table}[H]
    \centering
    \setlength\tabcolsep{2pt}
    \scriptsize
    \caption{Affordance Sheet, inspired by Model Cards~\cite{mitchell2019model}, to favour transparency and reproducibility of works for visual affordance predictions conditioned on robotic tasks. Example filled with VLTP~\cite{chen2024vltp} details.}
    \begin{tabular}{|l |c c c c c c|}
    \hline
    \multicolumn{7}{|c|}{\textbf{VLTP}} \\
    \hline
    \textbf{Affordance task}
     & OBJL & FUNC & FUNS & EPE & EIS & \\
     & \bbox & \wbox & \wbox & \wbox & \wbox & \\
    \hline
    \parbox{0.28\columnwidth}{\textbf{Datasets\\(RC1)}}
    & \multicolumn{1}{l}{\textit{Name:}} & \multicolumn{5}{l|}{RIO} \\ 
    & \multicolumn{1}{l}{\textit{Record link*:}} & \multicolumn{5}{l|}{\parbox{0.4\columnwidth}{\url{https://drive.google.com/drive/folders/1IAvh8tBGS3WWgV4SbVoqhwCkmyoSFffh}}} \\
    & \multicolumn{1}{l}{\textit{Licence:}} & \multicolumn{5}{l|}{\parbox{0.4\columnwidth}{\strut - \strut}}\\  
    \hline
    \parbox{0.28\columnwidth}{\textbf{Proposed method \\ (RC2, RC3)}} & \multicolumn{1}{l}{\textit{Record link*:}} & \multicolumn{5}{l|}{\parbox{0.4\columnwidth}{\url{-}}} \\ 
    & \multicolumn{1}{l}{\textit{Code link:}} & \multicolumn{5}{l|}{\parbox{0.4\columnwidth}{\url{https://github.com/HanningChen/VLTP/tree/main}}} \\
    & \multicolumn{1}{l}{ \textit{Model card}:} & \multicolumn{5}{l|}{\wbox} \\
    & \multicolumn{1}{l}{\textit{Licence:}} & \multicolumn{5}{l|}{Apache 2.0} \\  
    \hline
    \parbox{0.28\columnwidth}{\textbf{Experimental setup\\ (RC4)}} & \multicolumn{6}{l|}{\textit{Data splits:}} \\
    & &  \multicolumn{5}{l|}{ 
    \begin{tabular}{l c}
    \toprule
    Set & Images \\
    \midrule
    Training & 27,696 \\
    Validation & - \\
    Testing & 17,218 \\
    \bottomrule
    \end{tabular}}
    \\
    & \multicolumn{6}{l|}{\textit{Hyperparameters:}} \\
    & & \multicolumn{5}{l|}{
    \begin{tabular}{l c}
    \toprule
    Name & Value \\
    \midrule
    batch size & - \\
    learning rate & - \\
    schedule & - \\
    patience & - \\ 
    optmizer & - \\
    momentum & -  \\ 
    weight decay & - \\ 
    resize & - \\
    flip & - \\
    \bottomrule
    \end{tabular}
    }\\
    & \multicolumn{1}{l}{\textit{Resize procedure:}} & \multicolumn{5}{l|}{-}\\  
    \hline
    \parbox{0.28\columnwidth}{\textbf{Performance measures\\(RC5)}} & \multicolumn{1}{l}{\textit{Description:}} & \multicolumn{5}{l|}{\parbox{0.4\columnwidth}{\strut Mean Intersection over Union (mIoU) evaluates how well a model’s predicted segmentation aligns with the ground truth segmentation by calculating the overlap between the predicted and actual regions, averaged for the selected classes. \strut}}\\  
    & \multicolumn{1}{l}{\textit{Formulation:}} & \multicolumn{5}{l|}{\parbox{0.4\linewidth}{
    $\frac{1}{O} \sum_{i=1}^O \frac{TP_i}{FP_i+FN_i+TP_i}$}}\\  
    & \multicolumn{1}{l}{\textit{Limitations:}} & \multicolumn{5}{l|}{\parbox{0.4\linewidth}{\strut mIoU does not take into account the similarity in shape between the predicted and annotated segmentation mask \strut}} \\    
    \hline
    \textbf{Robot validation} 
    & \multicolumn{1}{l}{\textit{Robot model:}} & \multicolumn{5}{l|}{\parbox{0.35\columnwidth}{-}}\\
    & \multicolumn{1}{l}{\textit{End-effector:}} & \multicolumn{5}{l|}{\parbox{0.35\columnwidth}{-}}\\
    & \multicolumn{1}{l}{\textit{Experiment:}} & \multicolumn{5}{l|}{\parbox{0.35\columnwidth}{-}}\\
    \hline
    \multicolumn{7}{|l|}{\parbox{0.98\columnwidth}{\scriptsize{\strut \textit{Legend}: OBJL:~object localisation; FUNC:~functional classification; FUNS:~functional segmentation; EPE:~hand pose estimation; EIS:~hand interaction synthesis;
    RC:~reproducibility challenge; '-': information not available. \\
    \textbf{Notes}: *data and weights of the trained model are recommended to be placed in a repository that favours long-term persistence and accessibility.
    \strut}}} \\
    \hline
    \end{tabular}
    \label{tab:affsheet}
    \vspace{-5pt}
\end{table}

\begin{table}[H]
    \centering
    \setlength\tabcolsep{2pt}
    \scriptsize
    \caption{Affordance Sheet, inspired by Model Cards~\cite{mitchell2019model}, to favour transparency and reproducibility of works for visual affordance predictions conditioned on robotic tasks. Example filled with EgoTopo~\cite{nagarajan2020ego} details.}
    \begin{tabular}{|l |c c c c c c|}
    \hline
    \multicolumn{7}{|c|}{\textbf{EgoTopo}} \\
    \hline
    \textbf{Affordance task}
     & OBJL & FUNC & FUNS & EPE & EIS & \\
     & \wbox & \bbox & \wbox & \wbox & \wbox & \\
    \hline
    \parbox{0.28\columnwidth}{\textbf{Datasets\\(RC1)}}
    & \multicolumn{1}{l}{\textit{Name:}} & \multicolumn{5}{l|}{EPIC-Kitchens} \\ 
    & \multicolumn{1}{l}{\textit{Record link*:}} & \multicolumn{5}{l|}{\parbox{0.4\columnwidth}{\url{https://data.bris.ac.uk/data/dataset/2g1n6qdydwa9u22shpxqzp0t8m}}} \\
    & \multicolumn{1}{l}{\textit{Licence:}} & \multicolumn{5}{l|}{\parbox{0.4\columnwidth}{\strut CC-BY-NC 4.0 \strut}}\\  
    \hline
    \parbox{0.28\columnwidth}{\textbf{Proposed method \\ (RC2, RC3)}} & \multicolumn{1}{l}{\textit{Record link*:}} & \multicolumn{5}{l|}{\parbox{0.4\columnwidth}{\url{https://dl.fbaipublicfiles.com/ego-topo/anticipation/pretrained.zip}}} \\ 
    & \multicolumn{1}{l}{\textit{Code link:}} & \multicolumn{5}{l|}{\url{https://github.com/facebookresearch/ego-topo}} \\
    & \multicolumn{1}{l}{ \textit{Model card}:} & \multicolumn{5}{l|}{\wbox} \\
    & \multicolumn{1}{l}{\textit{Licence:}} & \multicolumn{5}{l|}{CC-BY-NC 4.0} \\  
    \hline
    \parbox{0.28\columnwidth}{\textbf{Experimental setup\\ (RC4)}} & \multicolumn{6}{l|}{\textit{Data splits:}} \\
    & &  \multicolumn{5}{l|}{ 
    \begin{tabular}{l c}
    \toprule
    Set & Images \\
    \midrule
    Training & - \\
    Validation & - \\
    Testing & 1,155 \\
    \bottomrule
    \end{tabular}}
    \\
    & \multicolumn{6}{l|}{\textit{Hyperparameters:}} \\
    & & \multicolumn{5}{l|}{
    \begin{tabular}{l c}
    \toprule
    Name & Value \\
    \midrule
    epochs & 20 \\
    batch size & 256 \\
    learning rate & 0.0001 \\
    schedule & 0.1x after 15 epochs \\
    patience & - \\ 
    optmizer & Adam \\
    momentum & -  \\ 
    weight decay & 0.000001 \\ 
    resize & - \\
    flip & - \\
    \bottomrule
    \end{tabular}
    }\\
    & \multicolumn{1}{l}{\textit{Resize procedure:}} & \multicolumn{5}{l|}{-}\\  
    \hline
    \parbox{0.28\columnwidth}{\textbf{Performance measures\\(RC5)}} & \multicolumn{1}{l}{\textit{Description:}} & \multicolumn{5}{l|}{\parbox{0.4\columnwidth}{\strut Mean  average precision (mAP) over all afforded interactions. \strut}}\\  
    & \multicolumn{1}{l}{\textit{Formulation:}} & \multicolumn{5}{l|}{\parbox{0.4\linewidth}{
    -}}\\  
    & \multicolumn{1}{l}{\textit{Limitations:}} & \multicolumn{5}{l|}{\parbox{0.4\linewidth}{\strut mAP score weights equally the classes, regardless of their frequency. \strut}} \\    
    \hline
    \textbf{Robot validation} 
    & \multicolumn{1}{l}{\textit{Robot model:}} & \multicolumn{5}{l|}{\parbox{0.35\columnwidth}{-}}\\
    & \multicolumn{1}{l}{\textit{End-effector:}} & \multicolumn{5}{l|}{\parbox{0.35\columnwidth}{-}}\\
    & \multicolumn{1}{l}{\textit{Experiment:}} & \multicolumn{5}{l|}{\parbox{0.35\columnwidth}{-}}\\
    \hline
    \multicolumn{7}{|l|}{\parbox{0.98\columnwidth}{\scriptsize{\strut \textit{Legend}: OBJL:~object localisation; FUNC:~functional classification; FUNS:~functional segmentation; EPE:~hand pose estimation; EIS:~hand interaction synthesis;
    RC:~reproducibility challenge; '-': information not available.\\
    \textbf{Notes}: *data and weights of the trained model are recommended to be placed in a repository that favours long-term persistence and accessibility.
    \strut}}} \\
    \hline
    \end{tabular}
    \label{tab:affsheet}
    \vspace{-5pt}
\end{table}

\end{document}